\documentclass{article}
\usepackage{fullpage}
\usepackage{cite}
\usepackage{graphicx}
\usepackage{amsmath,amsthm}
\usepackage{amssymb}
\usepackage{booktabs}
\usepackage{enumitem}
\usepackage[algo2e]{algorithm2e}
\usepackage{algorithm}
\usepackage{algpseudocode}
\usepackage{appendix}
\newtheorem{prop}{Proposition}
\newtheorem{thm}{Theorem}
\newtheorem{cor}{Corollary}
\newtheorem{definition}{Definition}

\newcommand{\anthon}{\textcolor{red}}
\newcommand{\farzan}{\textcolor{blue}}

\usepackage[pagebackref,breaklinks,colorlinks]{hyperref}

\allowdisplaybreaks
\usepackage[capitalize]{cleveref}
\crefname{section}{Sec.}{Secs.}
\Crefname{section}{Section}{Sections}
\Crefname{table}{Table}{Tables}
\crefname{table}{Tab.}{Tabs.}

\title{MoreauGrad: Sparse and Robust Interpretation of Neural Networks via Moreau Envelope}

\date{}

\author{
Jingwei~Zhang\thanks{Department of Computer Science and Engineering, The Chinese University of Hong Kong, jwzhang22@cse.cuhk.edu.hk}~, 
Farzan~Farnia\thanks{Department of Computer Science and Engineering, The Chinese University of Hong Kong, farnia@cse.cuhk.edu.hk}
	}

\begin{document}


\maketitle

%%%%%%%%% ABSTRACT
\begin{abstract}
   Explaining the predictions of deep neural nets has been a topic of great interest in the computer vision literature. While several gradient-based interpretation schemes have been proposed to reveal the influential variables in a neural net's prediction, standard gradient-based interpretation frameworks have been commonly observed to lack robustness to input perturbations and flexibility for incorporating prior knowledge of sparsity and group-sparsity structures. In this work, we propose MoreauGrad\footnote{The paper's code is available at \url{https://github.com/buyeah1109/MoreauGrad}} as an interpretation scheme based on the classifier neural net's Moreau envelope. We demonstrate that MoreauGrad results in a smooth and robust interpretation of a multi-layer neural network and can be efficiently computed through first-order optimization methods. Furthermore, we show that MoreauGrad can be naturally combined with $L_1$-norm regularization techniques to output a sparse or group-sparse explanation which are prior conditions applicable to a wide range of  deep learning applications. We empirically evaluate the proposed MoreauGrad scheme on standard computer vision datasets, showing the qualitative and quantitative success of the MoreauGrad approach in comparison to standard gradient-based interpretation methods.
\end{abstract}

%%%%%%%%% BODY TEXT
\section{Introduction}
% \label{sec:intro}

Deep neural networks (DNNs) have achieved state-of-the-art performance in many computer vision problems including image classification \cite{krizhevsky2017imagenet}, object detection \cite{zhao2019object}, and medical image analysis \cite{shen2017deep}. While they manage to attain super-human scores on standard image and speech recognition tasks, a reliable application of deep learning models to real-world problems requires an interpretation of their predictions to help domain experts understand and investigate the basis of their predictions. Over the past few years, developing and analyzing interpretation schemes that reveal the influential features in a neural network's prediction have attracted great interest in the computer vision community. 

A standard approach for interpreting neural nets' predictions is to analyze the gradient of their prediction score function at or around an input data point. Such gradient-based interpretation mechanisms result in a feature saliency map revealing the influential variables that locally affect the neural net's assigned prediction score. Three well-known examples of gradient-based interpretation schemes are the simple gradient \cite{simonyan2013deep}, integrated gradients \cite{sundararajan2017axiomatic}, and DeepLIFT \cite{shrikumar2017learning} methods. While the mentioned methods have found many applications in explaining neural nets' predictions, they have been observed to lack robustness to input perturbations and to output a dense noisy saliency map in their application to computer vision datasets \cite{ghorbani2019interpretation,heo2019fooling}. Consequently, these gradient-based explanations can be considerably altered by minor random or adversarial input noise.

A widely-used approach to improve the robustness and sharpness of gradient-based interpretations is SmoothGrad \cite{smilkov2017smoothgrad} which applies Gaussian smoothing to the mentioned gradient-based interpretation methods. As shown by \cite{smilkov2017smoothgrad}, SmoothGrad can significantly boost the visual quality of a neural net's gradient-based saliency map. On the other hand, SmoothGrad typically leads to a dense interpretation vector and remains inflexible to incorporate prior knowledge of sparsity and group-sparsity structures. Since a sparse saliency map is an applicable assumption to several image classification problems where a relatively small group of input variables can completely determine the image label,  a counterpart of SmoothGrad which can simultaneously achieve sparse and robust interpretation will be useful in computer vision applications.    

\begin{figure}
    \centering
    \includegraphics[width=0.9325\linewidth]{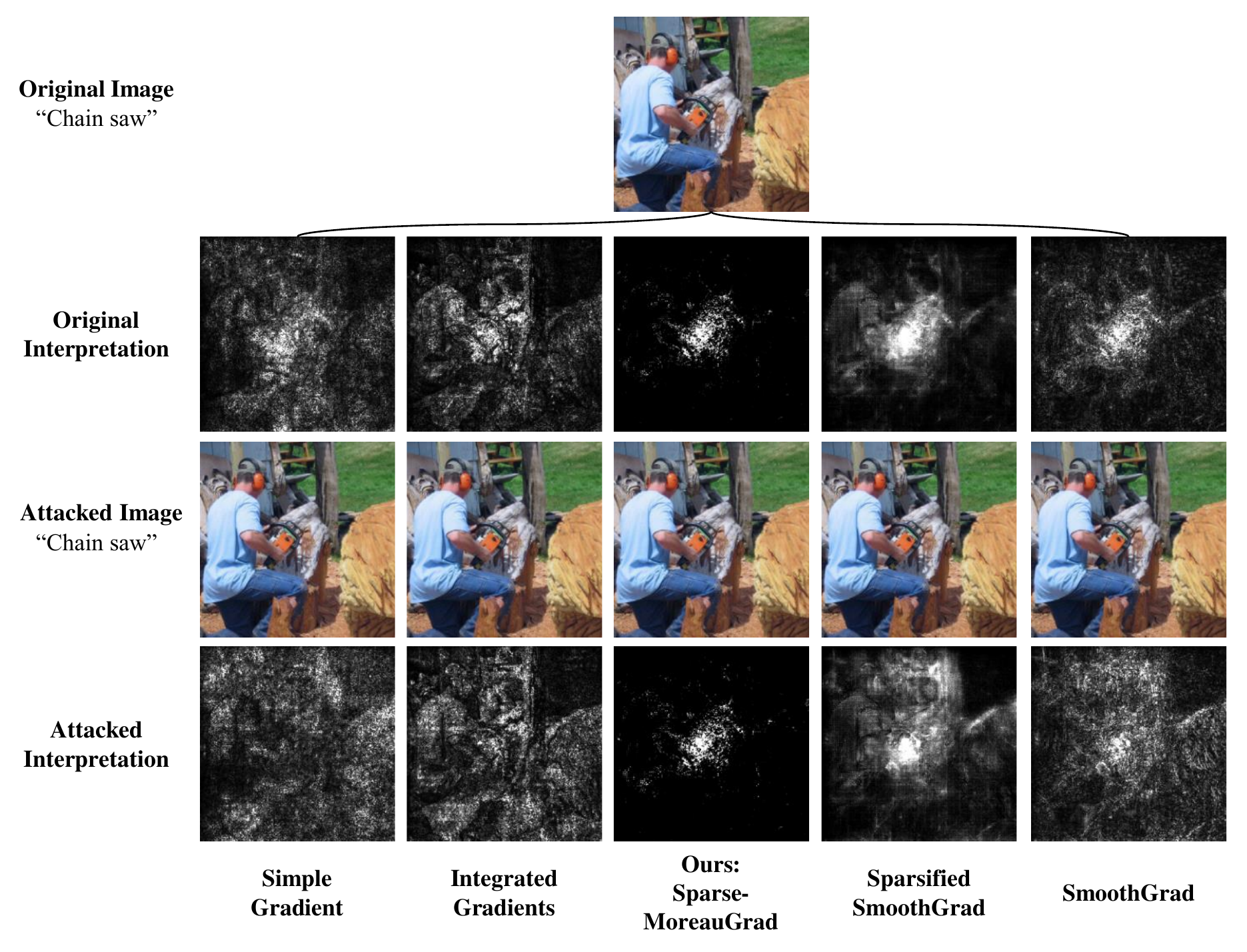}
    \caption{Interpretation of Sparse MoreauGrad (ours) vs. standard gradient-based baselines on an ImageNet sample before and after adding a norm-bounded interpretation adversarial perturbation.}\vspace{-3mm}
    \label{fig:intro}
\end{figure}

In this paper, we propose a novel approach, which we call \emph{MoreauGrad}, to achieve a provably smooth gradient-based interpretation with potential sparsity or group-sparsity properties. The proposed MoreauGrad outputs the gradient of a classifier's Moreau envelope which is a useful optimization tool for enforcing smoothness in a target function. %As we discuss in later sections, MoreauGrad searches for a direction in the proximity of the original sample with the most alteration in the prediction score function. 
We leverage convex analysis to show that MoreauGrad behaves smoothly around an input sample and therefore provides an alternative optimization-based approach to SmoothGrad for achieving a smoothly-changing saliency map. As a result, we demonstrate that similar to SmoothGrad, MoreauGrad offers robustness to input perturbations, since a norm-bounded perturbation will only lead to a bounded change to the MoreauGrad interpretation.   

Next, we show that MoreauGrad can be flexibly combined with $L_1$-norm-based regularization penalties to output sparse and group-sparse interpretations. Our proposed combinations, Sparse MoreauGrad and Group-Sparse MoreauGrad, take advantage of elastic-net \cite{zou2005regularization} and group-norm \cite{meier2008group} penalty terms to enforce sparse and group-sparse saliency maps, respectively. We show that these extensions of MoreauGrad preserve the smoothness and robustness properties of the original MoreauGrad scheme. Therefore, our discussion demonstrates the adaptable nature of MoreauGrad for incorporating prior knowledge of sparsity structures in the output interpretation.

%Additionally, we discuss the optimization of the interpretation vector in the MoreauGrad approach. We show that under mild regularity assumptions on the smoothness coefficient of the neural net classifier, the MoreauGrad interpretation can be efficiently computed via standard gradient-based optimization methods. In the case of vanilla MoreauGrad, we show that the standard or accelerated gradient descent algorithms can efficiently solve the MoreauGrad optimization with global convergence guarantees. For the $L_1$-norm-based extensions of MoreauGrad, we prove that standard or accelerated proximal gradient descent methods are capable of solving the optimization problem with global convergence guarantees. The proposed algorithms extend the well-known Iterative Iterative Shrinkage-Thresholding Algorithm (ISTA) and Fast Iterative Shrinkage-Thresholding Algorithm (FISTA) methods from the LASSO and Group-LASSO problems to the neural-net-based optimization problem in MoreauGrad.

Finally, we present the empirical results of our numerical experiments applying MoreauGrad to standard image recognition datasets and neural net architectures. We compare the numerical performance of MoreauGrad with standard gradient-based interpretation baselines. Our numerical results indicate the satisfactory performance of vanilla and $L_1$-norm-based MoreauGrad in terms of visual quality and robustness. Figure~\ref{fig:intro} shows the robustness and sparsity of the Sparse MoreauGrad interpretation applied to an ImageNet sample in comparison to standard gradient-based saliency maps. As this and our other empirical findings suggest, MoreauGrad can outperform standard baselines in terms of the sparsity and robustness properties of the output interpretation. In the following, we summarize the main contributions of this paper:

\begin{itemize}[leftmargin=*]
\item Proposing MoreauGrad as an interpretation scheme based on a classifier function's Moreau envelope
\item Analyzing the smoothness and robustness properties of MoreauGrad by leveraging convex analysis
\item Introducing $L_1$-regularized Sparse MoreauGrad to obtain an interpretation satisfying prior sparsity  conditions
\item Providing numerical results supporting MoreauGrad over standard  image recognition datasets

\end{itemize}
%-------------------------------------------------------------------------

% Update the cvpr.cls to do the following automatically.
% For this citation style, keep multiple citations in numerical (not
% chronological) order, so prefer \cite{Alpher03,Alpher02,Authors14} to
% \cite{Alpher02,Alpher03,Authors14}.

\section{Related Work}
\noindent \textbf{Gradient-based Interpretation.} 
A large body of related works develop gradient-based interpretation methods.
Simonyan et al. \cite{simonyan2013deep} propose to calculate the gradient of a classifier's output with respect to an input image. %In principal, the absolute value of pixel intensities in the class-specific saliency map denote how important the corresponding input pixels to the decision of a specific class by a neural network.
The simple gradient approach in \cite{simonyan2013deep} has been improved  by several related works. Notably, the method of Integrated Gradients \cite{sundararajan2017axiomatic} is capable of keeping highly relevant pixels in the saliency map by aggregating gradients of image samples. SmoothGrad \cite{smilkov2017smoothgrad} removes noise in saliency maps by adding Gaussian-random noise to the input image. The CAM method \cite{zhou2016learning} analyzes the information from global average pooling layer for localization, and  
Grad-CAM++ \cite{chattopadhay2018grad} improves over Grad-CAM \cite{selvaraju2017grad} and generates coarse heat-maps with improved multi-object localization. The NormGrad \cite{rebuffi2020there} focuses on the weight-based gradient to analyze the contribution of each image region. DeepLIFT \cite{shrikumar2017learning} uses difference from reference to propagate an attribution signal. However, the mentioned gradient-based methods do not obtain a sparse interpretation, and their proper combination with $L_1$-regularization to promote sparsity remains highly non-trivial and challenging. On the other hand, our proposed MoreauGrad can be smoothly equipped with $L_1$-regularization to output sparse interpretations and can further capture group-sparsity structures.

\noindent \textbf{Mask-based Interpretation.}
Mask-based interpretation methods rely on adversarial perturbations to interpret neural nets. Applying a mask which perturbs the neural net input, the importance of input pixels is measured by a masked-based method. %Mask-based interpretation usually plays a "deletion game" or "preservation game", depending on whether the mask is used for deleting or preserving the image region. 
This approach to explaining neural nets has been successfully applied in References \cite{wagner2019interpretable, fong2017interpretable,lim2021building, dabkowski2017real} and has been shown to benefit from dynamic perturbations \cite{ivanovs2021perturbation}. More specifically, Dabkowski and Gal \cite{dabkowski2017real} introduce a real-time mask-based detection method; Fong and Vedaldi \cite{fong2017interpretable} develop a model-agnostic approach with interpretable perturbations; Wagner et al. \cite{wagner2019interpretable} propose a method that could generate fine-grained visual interpretations. Moreover, Lim et al. \cite{lim2021building} leverage local smoothness to enhance their robustness towards samples attacked by PGD \cite{madry2017towards}. However, \cite{fong2017interpretable} and \cite{dabkowski2017real} show that perturbation-based interpretation methods are still vulnerable to adversarial perturbations. %, and hence the existing trade-off between visual quality and robustness is a dilemma.

We note that the discussed methods depend on optimizing perturbation masks for interpretations, and due to the non-convex nature of neural net loss functions, their interpretation remains sensitive to input perturbations. In contrast, our proposed MoreauGrad can provably smooth the neural net score function, and can adapt to non-convex functions using norm regularization. Hence, MoreauGrad can improve both the sparsity and robustness of the interpretation.

%Specifically, although mask-based interpretation methods share some similarity with our proposed MoreauGrad in terms of the optimization, our method is not perturbation-based. \anthon{I modified some description here, please check whether it's well-stated. I also commented out some text here.}% Instead, we leverage the Moreau envelope to optimize an approximate input in order to smooth the neural network. The substantial properties of Moreau envelope enable the sparse interpretation, as well as the theoretical guarantee in robustness against interpretation attacks.

\noindent \textbf{Robust Interpretation.}
The robustness of interpretation methods has been a subject of great interest in the literature.
Ghorbani et al. \cite{ghorbani2019interpretation} introduce a gradient-based adversarial attack method to alter the neural nets' interpretation. %They introduced a $L_\infty$ interpretation attack method, which makes perceptively indistinguishable changes to input images but significantly influence the interpretation, without changing the classification results. This raises the trust concern of neural network interpretation, as for two visually identical images with the same classification results, the interpretations for these two images could be substantially different. 
Dombrowski  et al. \cite{dombrowski2019explanations} demonstrate that interpretations could be manipulated, and they suggest improving the robustness via smoothing the neural net classifier. Heo et al. \cite{heo2019fooling} propose a manipulation method that is capable of generalizing across datasets. Subramanya et al. \cite{subramanya2019fooling} create adversarial patches fooling both the classifier and the interpretation. %These studies suggest the fragility of the pixel-level interpretation. 

To improve the robustness, Sparsified-SmoothGrad  \cite{levine2019certifiably} combines a sparsification technique with Gaussian smoothing to %ensure their interpretation method is certifiably robust to interpretation attacks. Specifically, although sparsification is applied in Sparsified SmoothGrad for certifiable robustness, their method is not actually sparse due to the averaging process and random perturbations. 
achieve certifiable robustness. The related works \cite{wagner2019interpretable,xu2018structured,fong2017interpretable,lim2021building, dabkowski2017real} discuss the application of adversarial defense methods against classification-based attacks to interpret the prediction of neural net classifiers. We note that these papers' main focus is not on defense schemes against interpretation-based attacks. Specifically, \cite{wagner2019interpretable} filter gradients internally during backpropogation, and \cite{lim2021building} leverage local smoothness to integrate more samples. %Their methods manage to improve the robustness by introducing additional regularization \cite{dabkowski2017real,fong2017interpretable}, filtered-gradients in backpropagation \cite{wagner2019interpretable}, and using extra local samples \cite{lim2021building}. \anthon{So we dont discuss their shortcomings here? then it might be harder to tell the difference between ours and theirs} 
Unlike the mentioned papers, our work proposes a model-agnostic optimization-based method which is capable of generating simultaneously sparse and robust interpretations.

\iffalse\anthon{model-agnostic is a key difference between ours and wagner's, they could generate sparse interpertation, and they discuss defense against classification attack} \farzan{what do you mean by model agnostic?} \anthon{black box, we dont need to know the structure of network} \farzan{but their method is applicable to all neural nets?} \anthon{at least same as ours, all CNN i suppose, they just modify the back-propagation process}. %We empirically compare the visual quality and robustness of our framework vs.  as well as empirical robustness results in the later sections.
\fi
%The papers \cite{dabkowski2017real, fong2017interpretable} compromise on visual quality like resolution, \cite{wagner2019interpretable} requires modification of networks, \cite{levine2019certifiably} generates fuzzy interpretations, \cite{lim2021building} is possible to bring in noise when local smoothness is not satisfied. In contrast to these methods, our proposed Sparse MoreauGrad is model-agnostic, capable to generate fine-grained, sparse and provably robust interpretations. We present the outstanding visual quality as well as empirical robustness results in the later sections.

\section{Preliminaries}
In this section, we review three standard interpretation methods as well as the notation and definitions in the paper.
\subsection{Notation and Definitions}
In the paper, we use notation $\mathbf{X}\in\mathbb{R}^d$ to denote the feature vector and $Y\in\{1,\ldots ,k\}$ to denote the label of a sample. In addition, $f_{\mathbf{w}}:\mathbb{R}^d\rightarrow \mathbb{R}^k$ denotes a neural net classifier with its weights contained in vector $\mathbf{w}\in\mathcal{W}$ where $\mathcal{W}$ is the feasible set of the neural net's weights. Here $f_{\mathbf{w}}$ maps the $d$-dimensional input $\mathbf{x}$ to a $k$-dimensional prediction vector containing the likelihood of each of the $k$ classes in the classification problem. For every class $c\in\{1,\ldots ,k\}$, we use the notation $f_{\mathbf{w},c}:\mathbb{R}^d\rightarrow \mathbb{R}$ to denote the $c$-th entry of $f_{\mathbf{w}}$'s output which corresponds to class $c$.

We use $\Vert\mathbf{x}\Vert_p$ to denote the $\ell_p$-norm of input vector $\mathbf{x}$. %Note that we define the $\ell_p$-operator norm of matrices as:
%\begin{equation}
%    \Vert A \Vert_p := \max_{\mathbf{x}\in\mathbb{R}^n:\Vert \mathbf{x}\Vert_p =  1}\: \big\Vert A\mathbf{x}\big\Vert_p.
%    \label{eq:label}
%\end{equation} 
Furthermore, we use notation $\Vert \mathbf{x} \Vert_{p,q}$ to denote the $\ell_{p,q}$-group-norm of $\mathbf{x}$ defined in the following equation for given variable subsets $S_1,\ldots ,S_t\subseteq\{1,\ldots,d\}$:
\begin{equation}
    \Vert \mathbf{x}\Vert_{p,q} = \left\Vert\,\bigl[ \Vert \mathbf{x}_{S_1}\Vert_p,\ldots , \Vert\mathbf{x}_{S_t} \Vert_p\bigr] \, \right\Vert_q
\end{equation}
In other words, $\Vert \mathbf{x} \Vert_{p,q}$ is the $\ell_q$-norm of a vector containing the $\ell_p$-norms of the subvectors of $\mathbf{x}$ characterized by index subsets $S_1,\ldots, S_t$.

\subsection{Gradient-based Saliency Maps}
In our theoretical and numerical analysis, we consider the following widely-used gradient-based interpretation baselines which apply to a classifier neural net $f_{\mathbf{w}}$ and predicted class $c$ for input $\mathbf{x}$:
\begin{enumerate}[leftmargin=*]
    \item  \textbf{Simple Gradient}: The simple gradient interpretation returns the saliency map of a neural net score function's gradient with respect to  input $\mathbf{x}$:
\begin{equation}
    \text{\rm SG}\bigl(f_{\mathbf{w},c},\mathbf{x}\bigr)\, := \, \nabla_{\mathbf{x}} f_{\mathbf{w},c}(\mathbf{x}).
    \label{eq:labelforthis}
\end{equation}
    In the applications of the simple gradient approach, $c$ is commonly chosen as the neural net's predicted label with the maximum prediction score. 
    \item \textbf{Integrated Gradients:} The integrated gradients approach approximates the integral of the neural net's gradient function between a reference point $\mathbf{x}^0$ and the input $\mathbf{x}$. Using $m$ intermediate points on the line segment connecting $\mathbf{x}^0$ and $\mathbf{x}$, the integrated gradient output will be
    \begin{equation}
    \text{\rm IG}\bigl(f_{\mathbf{w},c},\mathbf{x}\bigr)\, := \, \frac{\Delta \mathbf{x}}{m}\sum_{i=1}^m \nabla_{\mathbf{x}} f_{\mathbf{w},c}\bigl(\mathbf{x}^0 + \frac{i}{m}\Delta \mathbf{x}\bigr).
    \end{equation}
    In the above $\Delta \mathbf{x}:= \mathbf{x}-\mathbf{x}^0$ denotes the difference between the target and reference points $\mathbf{x},\mathbf{x}^0$.

    %\item \textbf{DeepLift:}
    
    \item \textbf{SmoothGrad:} SmoothGrad considers the averaged simple gradient score over an additive random perturbation $Z$ drawn according to an isotropic Gaussian distribution $Z\sim\mathcal{N}(\mathbf{0},\sigma^2 I_d)$. In practice, the SmoothGrad interpretation is estimated over a number $t$ of independently drawn noise vectors $\mathbf{z}_1,\ldots,\mathbf{z}_t\stackrel{\textrm{i.i.d.}}{\sim}\mathcal{N}(\mathbf{0},\sigma^2 I_d)$ according to the zero-mean Gaussian distribution:
    \begin{align}
        \text{\rm SmoothGrad}\bigl(f_{\mathbf{w},c},\mathbf{x}\bigr)\, :=& \;\mathbb{E}\bigl[ \nabla_{\mathbf{x}} f_{\mathbf{w},c}(\mathbf{x} + Z) \bigr] \;
        \approx \; \frac{1}{t}\sum_{i=1}^t \nabla_{\mathbf{x}} f_{\mathbf{w},c}(\mathbf{x} + \mathbf{z}_i). 
    \end{align}
   % Note that the accuracy of the above approximation can be improved by choosing a larger $m$; however, a greater $m$ leads to higher computational complexity for computing the gradient over a larger set of perturbed samples. 
\end{enumerate}

\section{MoreauGrad: An Optimization-based Interpretation Framework}
As discussed earlier, smooth classifier functions with a Lipschitz gradient help to obtain a robust explanation of neural nets. %The SmoothGrad approach ensures the smoothness property through randomized smoothing with a Gaussian noise. 
Here, we propose an optimization-based smoothing approach based on Moreau-Yosida regularization. To introduce this optimization-based approach, we first define a function's Moreau envelope.
\begin{definition}
Given regularization parameter $\rho>0$, we define the Moreau envelope of a function $g:\mathbb{R}^d\rightarrow \mathbb{R}$ as: 
\begin{equation}
    g^{\rho} (\mathbf{x}) \, :=\, \min_{\widetilde{\mathbf{x}}\in\mathbb{R}^d}\:g \bigl(\widetilde{\mathbf{x}}\bigr) +\frac{1}{2\rho}\bigl\Vert \widetilde{\mathbf{x}} - \mathbf{x}\bigr\Vert_2^2. 
    \label{eq:MoreauGrad}
\end{equation}
\end{definition}
\iffalse
\begin{equation}
    f^{\rho}_{\mathbf{w},c} (\mathbf{x}) \, :=\, \min_{\boldsymbol{\delta}\in\mathbb{R}^d}\:f_{\mathbf{w},c} (\mathbf{x} + \boldsymbol{\delta}) +\frac{1}{2\rho}\big\Vert \boldsymbol{\delta}\big\Vert_2^2. 
    \label{eq:MoreauGrad}
\end{equation}
\fi
In the above definition, $\rho>0$ represents the Moreau-Yosida regularization coefficient. Applying the Moreau envelope, we propose the MoreauGrad interpretation as the gradient of the classifier's Moreau envelope at an input $\mathbf{x}$.
\begin{definition}
    Given regularization parameter $\rho>0$, we define the MoreauGrad interpretation $\text{\rm MG}_{\rho}:\mathbb{R}^d\rightarrow\mathbb{R}^d$ of a neural net $f_{\mathbf{w}}$ predicting class $c$ for input $\mathbf{x}$ as 
    \begin{equation*}
        \text{\rm MG}_{\rho}(f_{\mathbf{w},c},\mathbf{x})\: := \: \nabla f^{\rho}_{\mathbf{w},c} (\mathbf{x}).
    \end{equation*}
\end{definition}
To compute and analyze the MoreauGrad explanation, we first discuss the optimization-based smoothing enforced by the Moreau envelope.
Note that the Moreau envelope is known as an optimization tool to turn non-smooth convex functions (e.g. $\ell_1$-norm) into smooth functions. Here, we discuss an extension of this result to weakly-convex functions which also apply to non-convex functions.
\begin{definition}
A function $g:\mathbb{R}^d\rightarrow \mathbb{R}$ is called $\lambda$-weakly convex if $\Phi(\mathbf{x}):= g(\mathbf{x})+\frac{\lambda}{2}\Vert \mathbf{x}\Vert_2^2$ is a convex function, i.e. for every $\mathbf{x}_1,\mathbf{x}_2\in\mathbb{R}^d$ and $0\le \alpha\le 1 $ we have:
\begin{align*}
    g\bigl(\alpha \mathbf{x}_1 + (1-\alpha)\mathbf{x}_2\bigr) \; \le \;  \alpha g(\mathbf{x}_1) + (1-\alpha)g(\mathbf{x}_2) + \frac{\lambda \alpha(1-\alpha) }{2} \bigl\Vert \mathbf{x}_1 - \mathbf{x}_2 \bigr\Vert^2_2.
\end{align*}
\end{definition}
%The following theorem shows the smoothness of the Moreau envelope of weakly-convex functions.
\iffalse
\begin{prop}
Suppose that $g:\mathbb{R}^d\rightarrow \mathbb{R}$ is $\lambda$-weakly convex. Then, the Moreau envelope $g^\rho :\mathbb{R}^d\rightarrow \mathbb{R}$ with coefficient $0<\rho<\frac{1}{\lambda}$ will be $\bigl( \frac{1}{\rho} - {\lambda} \bigr)$-smooth, i.e. it satisfies the following inequality for every $\mathbf{x}_1 , \mathbf{x}_2$:
\begin{equation*}
    \bigl\Vert \nabla g^\rho(\mathbf{x}_1) - \nabla g^\rho(\mathbf{x}_2) \bigr\Vert_2 \, \le\, \bigl( \frac{1}{\rho} - {\lambda} \bigr) \bigl\Vert \mathbf{x}_1 - \mathbf{x}_2 \bigr\Vert_2.
\end{equation*}
\end{prop}
\begin{proof}
We defer the proof to the Appendix.
\end{proof}

\fi

\begin{thm}\label{Theorem: Moreau Envelope}
%Suppose that function $g:\mathbb{R}^d\rightarrow \mathbb{R}$ is $\lambda$-smooth, i.e. its gradient satisfies $\bigl\Vert \nabla g(\mathbf{x}_1) - \nabla g(\mathbf{x}_2) \bigr\Vert_2 \le \lambda \Vert \mathbf{x}_1- \mathbf{x}_2\Vert_2 $. 
Suppose that $g:\mathbb{R}^d\rightarrow \mathbb{R}$ is a $\lambda$-weakly convex function. Assuming that $0<\rho<\frac{1}{\lambda}$, the followings hold for the optimization problem of the  Moreau envelope $g^\rho$ and the optimal solution ${\widetilde{x}}^*_\rho(\mathbf{x})$ solving the optimization problem:
\begin{enumerate}[leftmargin=*,itemsep=0pt]
 \item The gradients of $g^{\rho}$ and $g$  are related as for every $\mathbf{x}$:
    \begin{equation*}
        \nabla g^\rho (\mathbf{x}) = \nabla g \bigl({\widetilde{x}}^*_\rho(\mathbf{x})\bigr).
    \end{equation*}
    \item The difference ${\widetilde{x}}^*_\rho(\mathbf{x})- \mathbf{x}$ is aligned with $g^{\rho}$'s gradient:
    \begin{equation*}
        \nabla g^\rho (\mathbf{x}) = \frac{-1}{\rho}\bigl(\, {\widetilde{x}}^*_\rho(\mathbf{x})- \mathbf{x}\, \bigr).
    \end{equation*}
    \item $g^\rho $ will be $ \max\{\frac{1}{\rho},\frac{\lambda}{1- \rho\lambda}\} $-smooth, i.e. for every $\mathbf{x}_1 , \mathbf{x}_2$:
\begin{equation*}
    \bigl\Vert \nabla g^\rho(\mathbf{x}_1) - \nabla g^\rho(\mathbf{x}_2) \bigr\Vert_2 \, \le\, \frac{1}{\min\bigl\{\rho,\frac{1}{\lambda}- \rho\bigr\}} \bigl\Vert \mathbf{x}_1 - \mathbf{x}_2 \bigr\Vert_2.
\end{equation*}

   % \item There exists a unique optimal solution ${\widetilde{x}}^*_\rho(\mathbf{x})$ satisfying the following equation
    %\begin{equation*}
    %    {\widetilde{x}}^*_\rho(\mathbf{x}) = \mathbf{x} + \rho \nabla g ({\widetilde{x}}^*_\rho(\mathbf{x}))
    %\end{equation*}

    %\item ${\widetilde{x}}^*_\rho(\mathbf{x})$ is a $\kappa=\bigl(\frac{1+\lambda\rho}{1-\lambda\rho}\bigr)$-Lipschitz function of $\mathbf{x}$, i.e. the following inequality holds for every $\mathbf{x}_1,\mathbf{x}_2$:
    %\begin{equation*}
    %    \bigl\Vert {\widetilde{x}}^*_\rho(\mathbf{x}_1) - {\widetilde{x}}^*_\rho(\mathbf{x}_2)\bigr\Vert_2\le \kappa \bigl\Vert \mathbf{x}_1 - \mathbf{x}_2\bigr\Vert_2.
    %\end{equation*}
\end{enumerate}
\end{thm}
\begin{proof}
    This theorem is known for convex functions. In the Appendix, we provide another proof for the result. 
\end{proof}
\begin{cor}\label{Corollary, Robustness}
Assume that the prediction score function $f_{\mathbf{w},c}:\mathbb{R}^d\rightarrow\mathbb{R}$ is $\lambda$-weakly convex. Then, the MoreauGrad interpretation ${\text{\rm MG}}_{\rho}$ will remain robust under an $\epsilon$-$\ell_2$-norm bounded perturbation $\Vert \boldsymbol{\delta}\Vert_2\le \epsilon$ as
\begin{equation*}
    \bigl\Vert {\text{\rm MG}}_{\rho}(\mathbf{x}+\boldsymbol{\delta}) - {\text{\rm MG}}_{\rho}(\mathbf{x}) \bigr\Vert_2 \le \frac{\epsilon}{\min\bigl\{\rho,\frac{1}{\lambda}- \rho\bigr\}}.
\end{equation*}
\end{cor}

The above results imply that by choosing a small enough coefficient $\rho$ the  Moreau envelope will be a differentiable smooth function. Moreover, the computation of the Moreau envelope will reduce to a convex optimization task that can be solved by standard or accelerated gradient descent with global convergence guarantees. Therefore, one can efficiently compute the MoreauGrad interpretation by solving the optimization problem via the gradient descent algorithm. Algorithm~\ref{alg: GraidentDescent MoreauGrad} applies gradient descent to compute the solution to the Moreau envelope optimization which according to Theorem \ref{Theorem: Moreau Envelope} yields the MoreauGrad explanation.

%As shown in the following corollary, the MoreauGrad explanation will enjoy robustness to norm-bounded input perturbations.

As discussed above, MoreauGrad will be provably robust as long as the regularization coefficient will dominate the weakly-convexity degree of the prediction score. In the following proposition, we show this condition can be enforced by applying either Gaussian smoothing.
\begin{prop}\label{Prop: Gaussian smoothing}
Suppose that $f_{\mathbf{w},c}$ is $L$-Lipschitz, that is for every $\mathbf{x}_1,\mathbf{x}_2$ $\vert f_{\mathbf{w},c}(\mathbf{x}_1)-f_{\mathbf{w},c}(\mathbf{x}_2)\vert \le L\Vert \mathbf{x}_2 - \mathbf{x}_1 \Vert_2$, but could be potentially non-differentiable and non-smooth. Then, $h_{\mathbf{w},c}(\mathbf{x}):=\mathbb{E}[f_{\mathbf{w},c}(\mathbf{x}+\mathbf{Z})]$ where $\mathbf{Z}\sim\mathcal{N}(\mathbf{0},\sigma^2I_{d\times d})$ will be $\frac{L\sqrt{d}}{\sigma}$-weakly convex. %, and for every $\mathbf{x}_1,\mathbf{x}_2$:
%\begin{equation*}
%   \big\Vert \nabla h_{\mathbf{w},c}(\mathbf{x}_2) - \nabla h_{\mathbf{w},c}(\mathbf{x}_1) \big\Vert_2 \le \frac{L\sqrt{d}}{\sigma} \bigl\Vert \mathbf{x}_2 - \mathbf{x}_1 \bigr\Vert_2 
%\end{equation*}
\end{prop}
\begin{proof}
    We postpone the proof to the Appendix.
\end{proof}
The above proposition suggests the regularized MoreauGrad which regularizes the neural net function to satisfy the weakly-convex condition through Gaussian smoothing.

\section{Sparse and Group-Sparse MoreauGrad}
To further extend the MoreauGrad approach to output sparsely-structured feature saliency maps, we further include an $L_1$-norm-based penalty term in the Moreau-Yosida regularization and define the following $L_1$-norm-based sparse and group-sparse Moreau envelope.

\begin{definition} \label{Definition: sparse-MoreauGrad}
    For a function $g:\mathbb{R}^d\rightarrow\mathbb{R}$ and regularization coefficients $\rho,\eta >0$, we define $L_1$-Moreau envelope $g_{L_1}^{\rho,\eta}$:
    \begin{equation*}
        g_{L_1}^{\rho,\eta}(\mathbf{x})\, := \, \min_{\widetilde{\mathbf{x}}\in\mathbb{R}^d}\: g(\widetilde{\mathbf{x}}) + \frac{1}{2\rho} \bigl\Vert\widetilde{\mathbf{x}}-\mathbf{x} \bigr\Vert^2_2 + {\eta} \bigl\Vert\widetilde{\mathbf{x}}-\mathbf{x} \bigr\Vert_1.
    \end{equation*}
    We also define $L_{2,1}$-Moreau envelope $g_{L_{2,1}}^{\rho,\eta}$ as
    \begin{equation*}
        g_{L_{2,1}}^{\rho,\eta}(\mathbf{x})\, := \, \min_{\widetilde{\mathbf{x}}\in\mathbb{R}^d}\: g(\widetilde{\mathbf{x}}) + \frac{1}{2\rho} \bigl\Vert\widetilde{\mathbf{x}}-\mathbf{x} \bigr\Vert^2_2 + {\eta} \bigl\Vert\widetilde{\mathbf{x}}-\mathbf{x} \bigr\Vert_{2,1}.
    \end{equation*}
    In the above, the group norm $\Vert \cdot\Vert_{2,1}$ is defined as  $ \Vert \mathbf{x}\Vert_{2,1} := \sum_{i=1}^t \Vert \mathbf{x}_{S_i}\Vert_2$ for given subsets $S_1,\ldots ,S_t \subseteq \{1,\ldots ,d\}$.
   
\end{definition}

\begin{definition}
    Given regularization coefficients $\rho,\eta>0$, we define the Sparse MoreauGrad ($\text{\rm S-MG}_{\rho,\eta}$) and Group-Sparse MoreauGrad ($\text{\rm GS-MG}_{\rho,\eta}$) interpretations as
    \begin{align*}
        \text{\rm S-MG}_{\rho,\eta}(f_{\mathbf{w},c},\mathbf{x})\, := &\, \frac{1}{\rho} \bigl(\,{\widetilde{\mathbf{x}}}_{L_1}^*(\mathbf{x}) - \mathbf{x}\, \bigr),\\ %\nabla f_{\mathbf{w},c,L_1}^{\rho,\eta}(\mathbf{x}),\\
        \;\; \text{\rm GS-MG}_{\rho,\eta}(f_{\mathbf{w},c},\mathbf{x})\, :=&\, \frac{1}{\rho} \bigl(\,{\widetilde{\mathbf{x}}}_{L_{2,1}}^*(\mathbf{x}) - \mathbf{x}\, \bigr),
    \end{align*}
where ${\widetilde{\mathbf{x}}}_{L_1}^*(\mathbf{x}) ,\, {\widetilde{\mathbf{x}}}_{L_{2,1}}^*(\mathbf{x}) $ denote the optimal solutions to the optimization tasks of $f_{\mathbf{w},c,L_1}^{\rho,\eta}(\mathbf{x}),\, f_{\mathbf{w},c,L_{2,1}}^{\rho,\eta}(\mathbf{x})$, respectively.
\end{definition}
In Theorem \ref{Thm: Sparse MoreauGrad}, we extend the shown results for the standard Moreau envelope to our proposed $L_1$-norm-based extensions of the Moreau envelope. Here, we use ${\text{\rm ST}}_{\alpha}$ and $ {\text{\rm GST}}_{\alpha}$ to denote sparse and group-sparse soft-thresholding functions defined entry-wise and group-entry-wise as
\begin{align*}
    {\text{\rm ST}}_{\alpha}(\mathbf{x})_i  := &\begin{cases}
    0\;\; &\text{\rm if}\; |x_i|\le \alpha \\
    x_i-\text{\rm sign}(x_i)\alpha\;\; &\text{\rm if}\; |x_i|> \alpha,
    \end{cases} \\
    {\text{\rm GST}}_{\alpha}(\mathbf{x})_{S_i}  := & \begin{cases}
    \mathbf{0}\;\; &\text{\rm if}\; \Vert \mathbf{x}_{S_i}\Vert_2\le \alpha \\
  \bigl( 1- \frac{\alpha}{\Vert \mathbf{x}_{S_i}\Vert_2} \bigr) \mathbf{x}_{S_i}\;\; &\text{\rm if}\; \Vert \mathbf{x}_{S_i}\Vert_2 > \alpha.
    \end{cases}
\end{align*}

\begin{thm}\label{Thm: Sparse MoreauGrad}
    Suppose that $g:\mathbb{R}^d\rightarrow\mathbb{R}$ is a $\lambda$-weakly convex function. Then, assuming that $0<\rho<\frac{1}{\lambda}$, Theorem \ref{Theorem: Moreau Envelope}'s parts 1 and 3 will further hold for the sparse Moreau envelope $g_{L_1}^{\rho,\eta}$ and group-sparse Moreau envelope $g_{L_{2,1}}^{\rho,\eta}$ and their optimization problems' optimal solutions $\widetilde{\mathbf{x}}^*_{\rho,\eta, L_1}(\mathbf{x})$ and $\widetilde{\mathbf{x}}^*_{\rho,\eta, L_{2,1}}(\mathbf{x})$. To parallel Theorem \ref{Theorem: Moreau Envelope}'s part 2 for $L_1$-Moreau envelope, the followings hold
    \begin{align*}
        {\text{\rm ST}}_{\rho\eta}\bigl(-\rho \nabla g^{\rho,\eta}_{L_{1}} (\mathbf{x})\bigr) \, &= \, \widetilde{\mathbf{x}}^*_{\rho,\eta, L_1}(\mathbf{x})- \mathbf{x} , \\
        {\text{\rm GST}}_{\rho\eta}\bigl(-\rho\nabla g^{\rho,\eta}_{L_{2,1}} (\mathbf{x})\bigr) \, &= \, \widetilde{\mathbf{x}}^*_{\rho,\eta, L_{2,1}}(\mathbf{x})- \mathbf{x}.
    \end{align*}   
\end{thm}
\begin{proof}
We defer the proof to the Appendix. 
\end{proof}
\begin{cor}
Suppose that the prediction score function $f_{\mathbf{w},c}$ is $\lambda$-weakly convex. Assuming that $0<\rho<\frac{1}{\lambda}$, the Sparse MoreauGrad $\text{\rm S-MG}_{\rho,\eta}$  and Group-Sparse MoreauGrad  $\text{\rm GS-MG}_{\rho,\eta}$ interpretations will be robust to every norm-bounded perturbation $\Vert \boldsymbol{\delta} \Vert_2\le \epsilon$ as:
\begin{align*}
    \bigl\Vert {\text{\rm S-MG}}_{\rho.\eta}(\mathbf{x}+\boldsymbol{\delta}) - {\text{\rm S-MG}}_{\rho,\eta}(\mathbf{x}) \bigr\Vert_2 \, & \le \, \frac{\epsilon}{\min\bigl\{\rho,\frac{1}{\lambda}- \rho\bigr\}}, \\
    \bigl\Vert {\text{\rm GS-MG}}_{\rho,\eta}(\mathbf{x}+\boldsymbol{\delta}) - {\text{\rm GS-MG}}_{\rho,\eta}(\mathbf{x}) \bigr\Vert_2 \, & \le \, \frac{\epsilon}{\min\bigl\{\rho,\frac{1}{\lambda}- \rho\bigr\}}.
\end{align*}\vspace{2mm}
\end{cor}

%Here, we decompose the objective function into two parts

%\section{Computation of MoreauGrad}
To compute the Sparse and Group-Sparse MoreauGrad, we  propose applying the proximal gradient descent algorithm as described in Algorithm~\ref{alg: GraidentDescent MoreauGrad}. Note that Algorithm~\ref{alg: GraidentDescent MoreauGrad} applies the soft-thresholding function as the proximal operator for the $L_1$-norm function present in Sparse MoreauGrad.

%\begin{figure}
\begin{algorithm}[H]
\textbf{Input}: data $\mathbf{x}$, label $c$, classifier $f_{\mathbf{w}}$, regulatization coeff. $\rho$, stepsize $\gamma$, noise std. parameter $\sigma$, number of updates $T$\\
\textbf{Initialize}
$\mathbf{x}^{(0)}=\mathbf{x}$, \\
\For{$t = 0, \ldots, T$}{
\uIf{Regularized Mode}{
    \textbf{Draw} noise vectors $\mathbf{z}_1,\ldots ,\mathbf{z}_m\sim\mathcal{N}(\mathbf{0},\sigma^2 I_{d\times d})$\vspace{.15cm}  \\
    \textbf{Compute} $\mathbf{g}_t = \frac{1}{m}\sum_{i=1}^m \nabla f_{\mathbf{w},c}(\mathbf{x}^{(t)} + \mathbf{z}_i)$
}
\Else{ \textbf{Compute} $\mathbf{g}_t = \nabla f_{\mathbf{w},c}(\mathbf{x}^{(t)})$ }
    \vspace{.15cm}
    \textbf{Update}  $\mathbf{x}^{(t+1)} \leftarrow (1-\frac{\gamma}{\rho})\mathbf{x}^{(t)} - \gamma(\mathbf{g}_t - \frac{1}{\rho}\mathbf{x})$\vspace{.15cm} \\
    \uIf{Sparse Mode}{
\textbf{Update} $\mathbf{x}^{(t+1)} \leftarrow \text{\rm SoftThreshold}_{{\gamma}{\eta}}\bigl(\mathbf{x}^{(t+1)} -\mathbf{x} \bigr) +\mathbf{x}$ 
    }{}
}
\textbf{Output} $\textrm{MG}(\mathbf{x})=\frac{1}{\rho}\bigl(\mathbf{x}^{(T)}-\mathbf{x}\bigr)$
\caption{MoreauGrad Interpretation}\label{alg: GraidentDescent MoreauGrad} 
\end{algorithm}
%\end{figure}

\begin{figure*}
    \centering
    \includegraphics[width=0.95\textwidth]{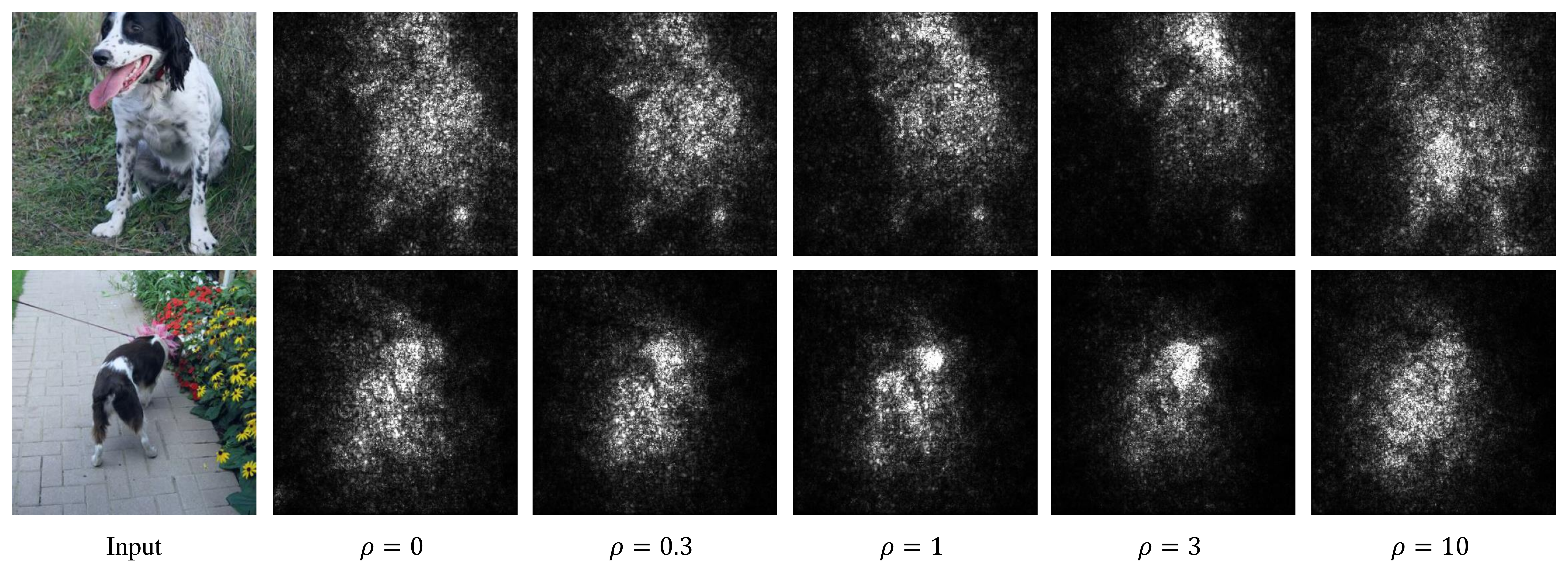}
    \caption{Visualization of MoreauGrad with various coefficient $\rho$'s. $\rho=0$ is Simple Gradient.% in this implementation. Both two rows denote the smoothed interpretation map $\nabla_x f_{\mathbf{w},c}(\mathbf{x})$.
    }
    \label{fig:lambda}\vspace{-3mm}
\end{figure*}

%For the sparse and group-sparse MoreauGrad, we propose a proximal gradient descent-based algorithm where we optimize the non-smooth norm penalty term via their proximal operator. 

\section{Numerical Results}

\begin{figure*}
    \centering
    \includegraphics[width=0.95\textwidth]{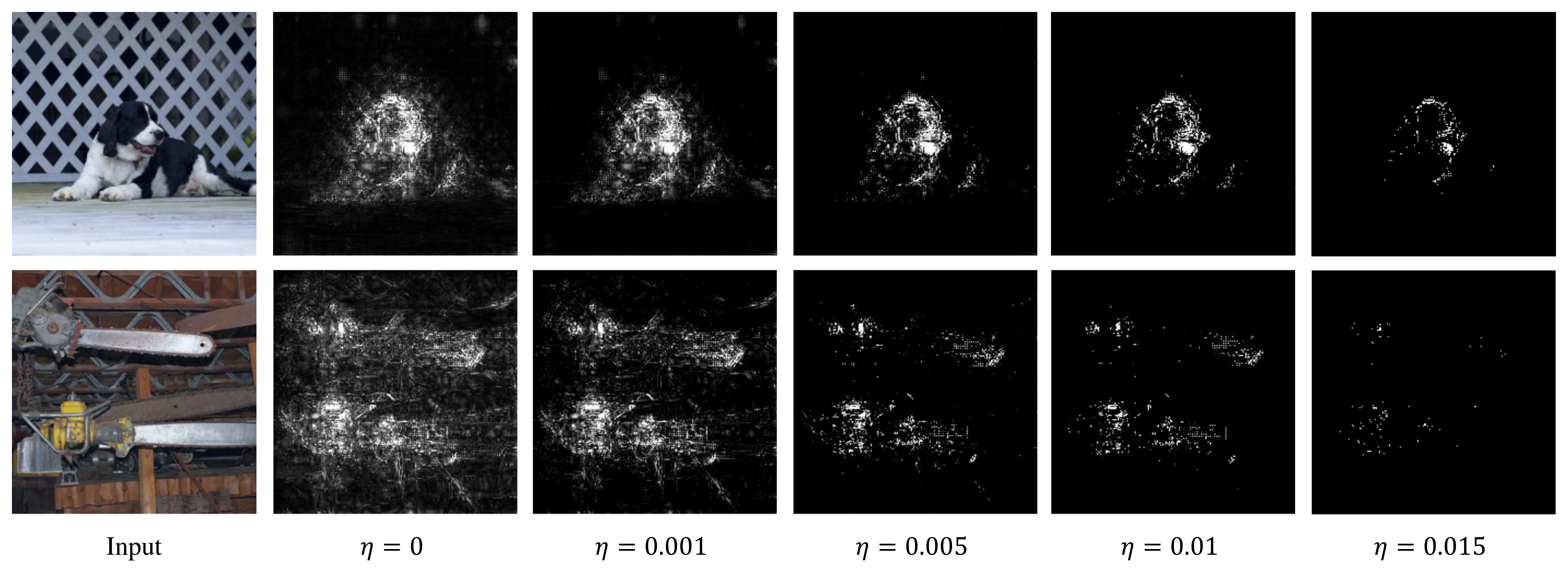}
    \caption{Visualization of Sparse MoreauGrad with various coefficient $\eta$'s. $\eta=0$ is Vanilla MoreauGrad.% These two rows focus on pixel-level sparsity with coefficient $\gamma_1$. $\gamma_1=0$ indicates the Vanilla MoreauGrad without $L_1$-regularization. As $\gamma_1$ increases, the interpretation maps become sparser in pixel-level. For Group-Sparse MoreauGrad, the effect is similar and the visualization is deferred to the Appendix.
    }\vspace{-3mm}
    \label{fig:l1}
\end{figure*}

We conduct several numerical experiments to evaluate the performance of the proposed MoreauGrad. Our designed experiments focus on the smoothness, sparsity, and robustness properties of  MoreauGrad interpretation maps as well as the feature maps of several standard baselines. In the following, we first describe the numerical setup in our experiments and then present the obtained numerical results on the qualitative and quantitative performance of interpretation methods.  

%In general, we introduce the effect of every parameter in MoreauGrad in \cref{fig:lambda} and \cref{fig:l1}, we visualize the qualitative comparison experiments in \cref{fig:compare}, and examine the robustness against interpretation attacks in \cref{fig:robust} and \cref{fig:line}.

\subsection{Experiment Setup}
In our numerical evaluation, we use the following standard image datasets: CIFAR-10 \cite{krizhevsky2009learning} consisting of 60,000 labeled samples with 10 different labels (50,000 training samples and 10,000 test samples), and ImageNet-1K \cite{deng2009imagenet} including 1.4 million labeled samples with 1,000 labels (10,000 test samples and 1.34 million training samples). For CIFAR-10 experiments, we trained a standard ResNet-18 \cite{he2016deep} neural network with the softplus activation. For ImageNet experiments, we used an EfficientNet-b0 network \cite{tan2019efficientnet} pre-trained on the ImageNet training data. In our experiments, we compared the MoreauGrad schemes with the following baselines: 1) the simple gradient \cite{simonyan2013deep}, 2) Integrated Gradients \cite{selvaraju2017grad}, 3) DeepLIFT \cite{shrikumar2017learning}, 4) SmoothGrad \cite{smilkov2017smoothgrad}, 5) Sparsified SmoothGrad \cite{levine2019certifiably}, 6) RelEx \cite{lim2021building}. We note that for baseline experiments we adopted the official implementations and conducted the experiments with hyperparameters suggested in their work. %We present the full implementation details in the Appendix \anthon{may remove for arxiv version:and note that our code is available in the supplementary material.} %at the anonymous Github repository  \url{https://github.com/buyeah1109/MoreauGrad}.

\subsection{Effect of Smoothness and Sparsity Parameters}
We ran the numerical experiments for unregularized Vanilla MoreauGrad with multiple smoothness coefficient $\rho$ values to show the effect of the Moreau envelope's regularization.
\Cref{fig:lambda} visualizes the effect of different $\rho$ on the Vanilla MoreauGrad saliency map. As can be seen in this figure, the saliency map qualitatively improves by increasing the value of $\rho$ from $0$ to $1$. Please note that for $\rho=0$, the MoreauGrad simplifies to the simple gradient interpretation.
However, as shown in Theorem \ref{Theorem: Moreau Envelope}
the proper performance of Vanilla MoreauGrad requires choosing a properly bounded $\rho$ value, which is consistent with our observation that when $\rho$ becomes too large, the Moreau envelope will be computationally difficult to optimize and the quality of interpretation maps could deteriorate to some extent. As numerically verified in both CIFAR-10 and ImageNet experiments, we used the rule of thumb $\rho=\frac{1}{\sqrt{\mathbb{E}[\Vert\mathbf{X}\Vert_2]}}$ measured over the empirical training data to set the value of $\rho$, which is equal to $1$ for the normalized samples in our experiments.

%\subsection{$L_1$ and Group Sparsity Penalty $\gamma_1 / \gamma_g$}

\begin{figure*}
    \centering
    \includegraphics[width=0.95\textwidth]{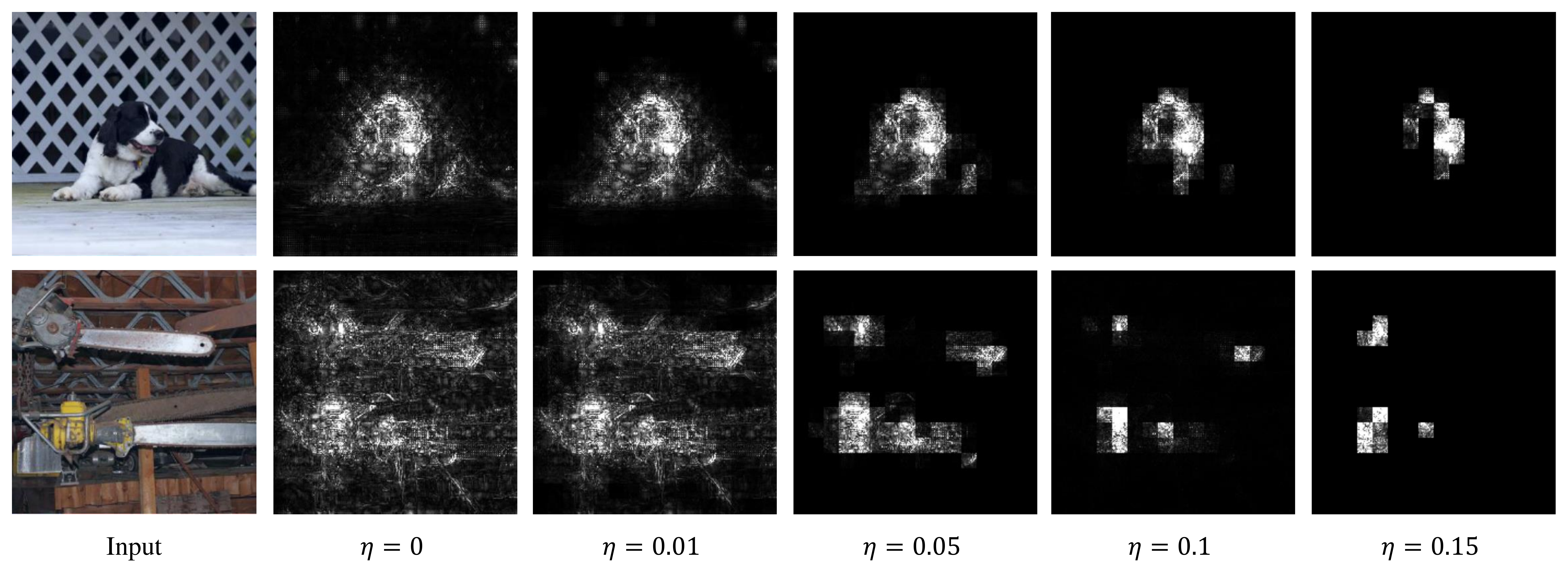}
    \caption{Visualization of Group-Sparse MoreauGrad maps with various coefficient $\eta$'s. %$\eta=0$ is Vanilla MoreauGrad.
    }\vspace{-3mm}
    \label{fig:groupnorm}
\end{figure*}

Regarding the sparsity hyperparameter $\eta$ in Sparse and Group-Sparse MoreauGrad experiments, we ran several experimental tests to properly tune the hyperparameter. Note that a greater coefficient $\eta$ enforces more strict sparsity or group-sparsity in the MoreauGrad interpretation,
%Flexibility in promoting sparsity is one of the distinguishable features of MoreauGrad. The proposed MoreauGrad could be naturally combined with $L_1$-regularization to promote sparsity and group-sparsity in interpretation maps, 
and the degree of sparsity could be simply adjusted by changing this coefficient $\eta$. As shown in \Cref{fig:l1}, in our experiments with different $\eta$ coefficients the interpretation map becomes sparser as we increase the $L_1$-norm penalty coefficient ${\eta}$. Similarly, to achieve a group-sparse interpretation, we used $L_{2,1}$-regularization on groups of adjacent pixels as discussed in Definition \ref{Definition: sparse-MoreauGrad}. The effect of the group-sparsity coefficient was similar to the sparse case in our experiments, as fewer pixel groups took non-zero values and the output interpretations showed more structured interpretation maps when choosing a larger coefficient ${\eta}$. %Due to limited space, 
The results with different group-sparsity hyperparameters are demonstrated in \Cref{fig:groupnorm}.

% \begin{figure}
%     \centering
%     \includegraphics[width=0.9\linewidth]{Figures/groupnorm.pdf}
%     \caption{Qualitative results with different GroupNorm coefficient $\gamma_g$. As $\gamma_g$ increases, the results become sparser and evolves into a more structured interpretation. The penalty $\lambda = 1$ in all cases.}
%     \label{fig:l1}
% \end{figure}

% \subsection{Relation between Saliency Map and $\delta^{*}$}

\subsection{Qualitative Comparison of MoreauGrad vs. Gradient-based Baselines}

\begin{figure}
    \centering
    \includegraphics[width=\textwidth]{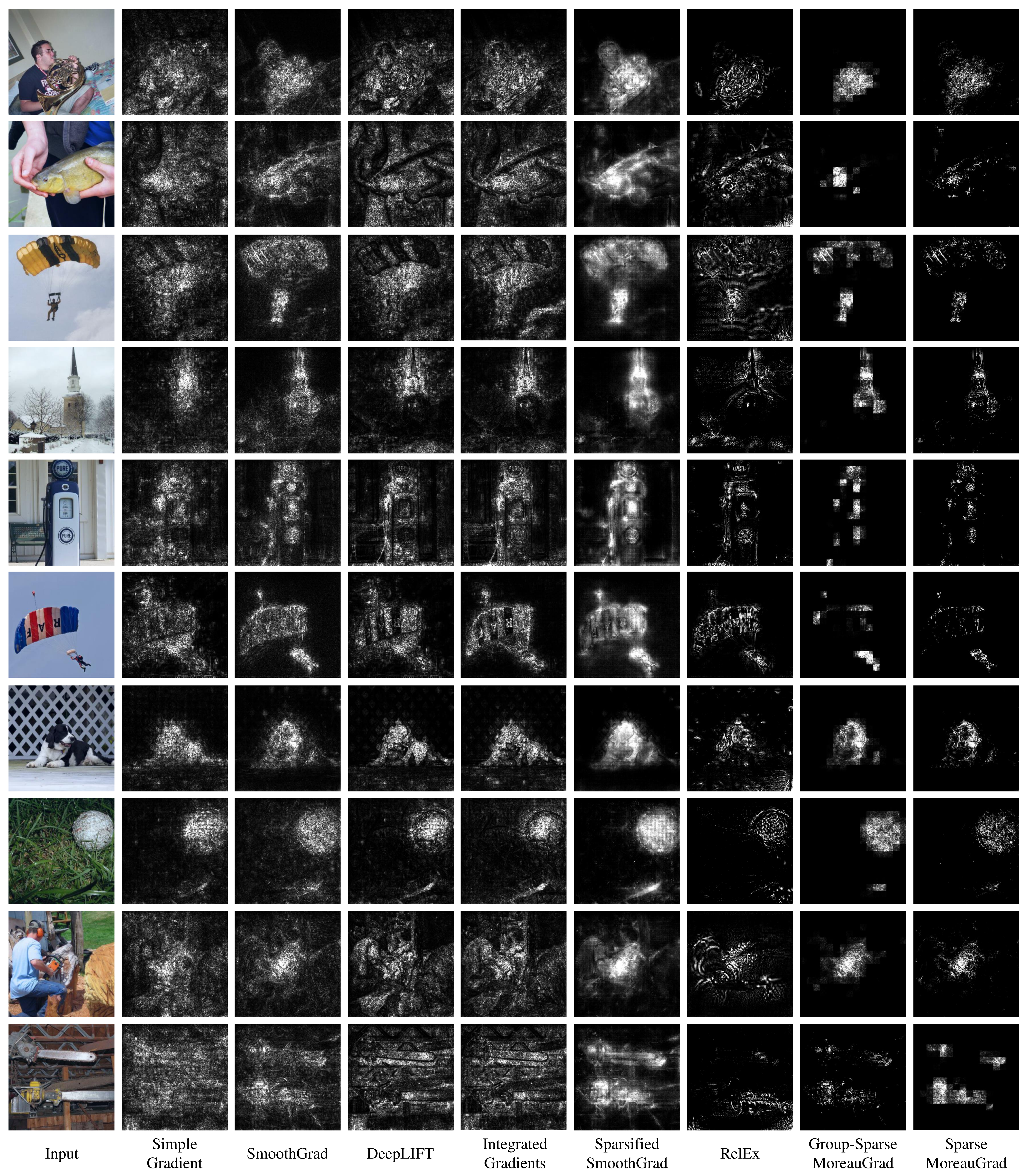}
    \caption{Qualitative comparison between Sparse, Group-Sparse MoreauGrad and the baselines. %The proposed Sparse and Group-Sparse MoreauGrad provide sharper interpretation maps. More examples are presented in the Appendix.
    }
    \label{fig:compare}
\end{figure}

In \Cref{fig:compare}, we illustrate the Sparse, and Group-Sparse MoreauGrad interpretation outputs as well as the saliency maps generated by the gradient-based baselines. %Precise implementation details are deferred to the Appendix for space efficiency.
The results demonstrate that MoreauGrad generates qualitatively sharp and, in the case of Sparse and Group-Sparse MoreauGrad, sparse interpretation maps. %Compared with other baselines, sparse interpretation maps could significantly help people concentrate on the most important pixel regions.
As shown in \Cref{fig:compare}, promoting sparsity in the MoreauGrad interpretation maps has improved the visual quality, and managed to erase the less relevant pixels like the background ones. Additionally, in the case of Group-Sparse MoreauGrad, the maps  exhibit both sparsity and connectivity of selected pixels.

\begin{figure}[t]
    \centering
\includegraphics[width=\textwidth]{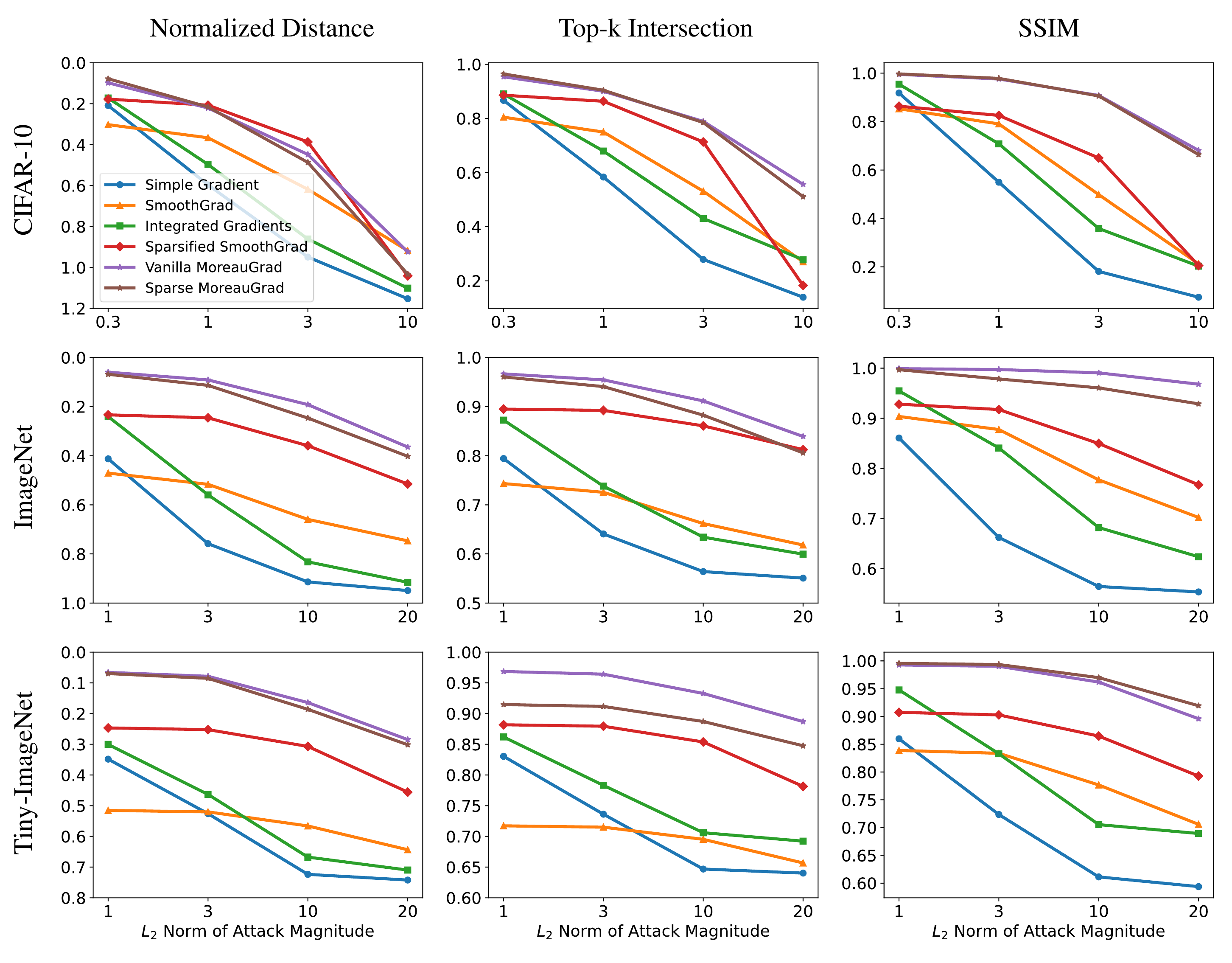}
    \caption{Quantitative robustness comparison between MoreauGrad and the baselines. %Higher data point indicates the interpretation method is more robust. Rows are robustness results of different datasets. Columns represent different robustness metrics. MoreauGrad is examined to have outstanding robustness against attacks in different image scales and metrics.
    } \vspace{-3mm}
    \label{fig:line}
\end{figure}

\subsection{Robustness}
%In this section, we illustrate that MoreauGrad could both preserve high visual quality and robustness, supported by visualization in \cref{fig:robust} and quantitative results in \cref{fig:line}.

\begin{figure*}
    \centering
    \includegraphics[width=0.9\textwidth]{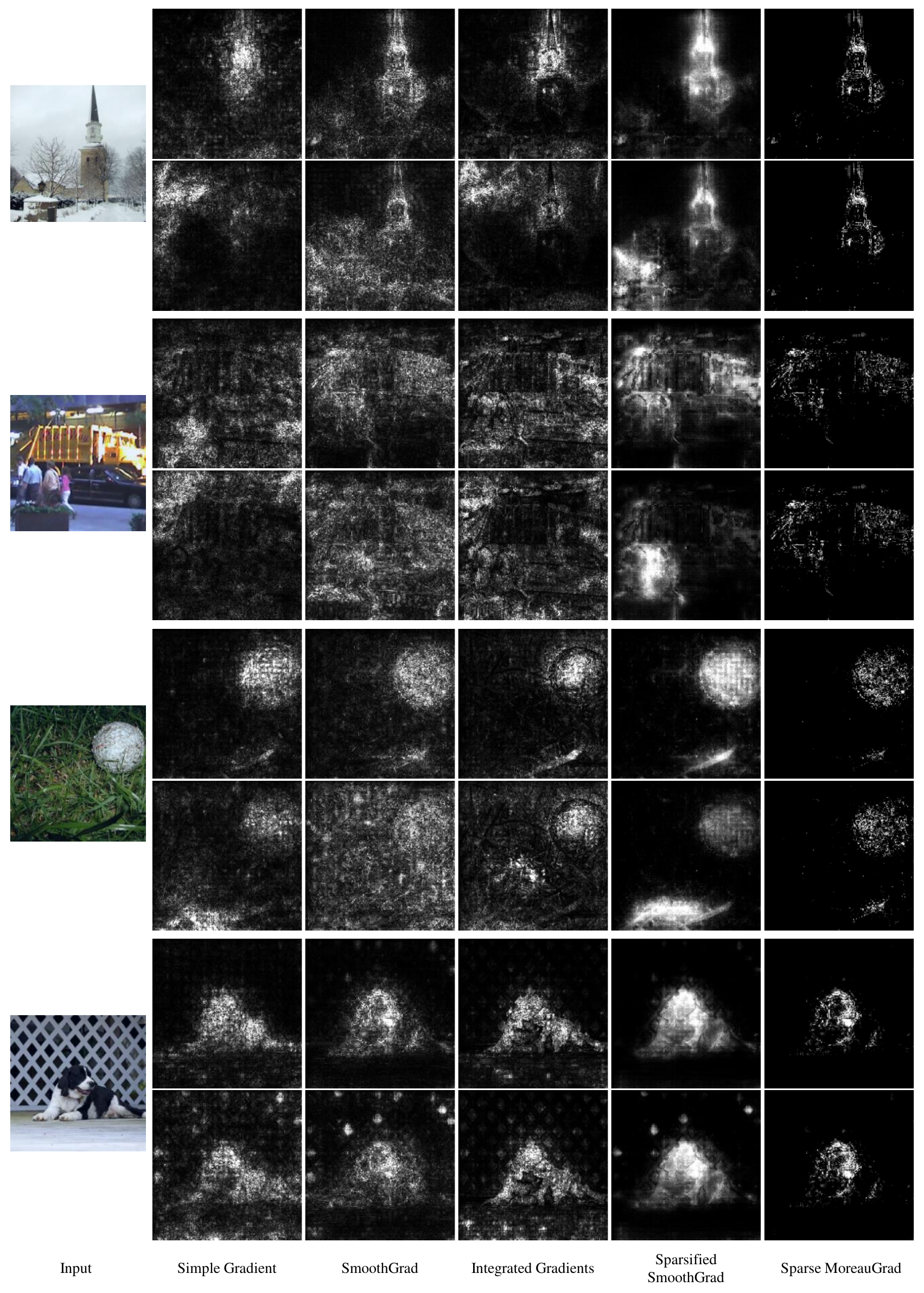}
    \caption{Visualization of robustness against interpretation attacks. %The left-most column demonstrates the input color image. 
    The top and bottom rows show original and attacked maps. %More examples could be found in the Appendix.
    \vspace{-2mm}
    }
    \label{fig:robust}
\end{figure*}

We qualitatively and quantitatively  evaluated the robustness of MoreauGrad interpretation. 
%\textbf{Interpretation attack method.}
To assess the empirical robustness of interpretation methods, we adopt a $L_2$-bounded interpretation attack method defined by \cite{levine2019certifiably}.  %Precise definition could be found in the Appendix and \cite{levine2019certifiably}. 
% \begin{equation}
%    
% \end{equation}
%
%\textbf{Evaluation metrics.}
Also, for quantifying the empirical robustness, we adopt three robustness metrics. The first metric is the Euclidean distance of the normalized interpretations before and after the attack:
\begin{equation}
    D(I(\mathbf{x}),I(\mathbf{x}'))= \big\Vert \frac{I(\mathbf{x})}{\Vert I(\mathbf{x}) \Vert_2} - \frac{I(\mathbf{x}')}{\Vert I(\mathbf{x}') \Vert_2} \big\Vert_2
    \label{eq:normalized_distance}
\end{equation}
%where $\frac{I}{\Vert I \Vert_2}$ denotes normalized original or perturbed interpretation map. 
Note that a larger distance between the normalized maps indicates a smaller similarity and a higher vulnerability of the interpretation method to adversarial attacks.

The second metric is the top-k intersection ratio. This metric is another standard robustness measure used in \cite{ghorbani2019interpretation, levine2019certifiably}. This metric measures the ratio of pixels that remain salient after the interpretation attack. A robust interpretation is expected to preserve most of the salient pixels under an attack. The third metric is the structural similarity index measure (SSIM) \cite{wang2004image}. A larger SSIM value indicates that the two input maps are more perceptively similar. %We defer the presentation of the full details of these metrics to the Appendix.

%\textbf{Robustness result and comparison.}
Using the above metrics, we compared the MoreauGrad schemes with the baseline methods. %Especially, Sparsified SmoothGrad \cite{levine2019certifiably} is provably robust to interpretation attack to a certain degree. 
As qualitatively shown in Figure~\ref{fig:robust}, using the same attack magnitude, the MoreauGrad interpretations are mostly similar before and after the norm-bounded attack. The qualitative robustness of MoreauGrad seems satisfactory compared to the baseline methods. Finally, Figure~\ref{fig:line} presents a  quantitative comparison of the robustness measures for the baselines and proposed MoreauGrad on CIFAR-10, 
  Tiny-ImageNet, and ImageNet datasets. As shown by these measures, MoreauGrad outperforms the baselines in terms of the  robustness metrics. %Note that the plots' x-axis indicates the interpretation attack $\ell_2$-norm. %, it specifies the allowed largest amount of pixel intensities that could be altered in the attack process. 

\vspace{-0.1cm}
\section{Conclusion}

In this work, we introduced MoreauGrad as an optimization-based interpretation method for deep neural networks. We demonstrated that MoreauGrad can be flexibly combined with $L_1$-regularization methods to output sparse and group-sparse interpretations. We further showed that the MoreauGrad output will enjoy robustness against input perturbations. While our analysis focuses on the sparsity and robustness of the MoreauGrad explanation, studying the consistency and transferability of MoreauGrad interpretations is an interesting future direction. Moreover, the application of MoreauGrad to convex and norm-regularized neural nets could be another topic for future study. Finally, our analysis of $\ell_1$-norm-based Moreau envelope could find independent applications to other deep learning problems. 
%%%%%%%%% REFERENCES

{
\bibliographystyle{unsrt}
\bibliography{ref}

\begin{thebibliography}{10}

\bibitem{krizhevsky2017imagenet}
Alex Krizhevsky, Ilya Sutskever, and Geoffrey~E Hinton.
\newblock Imagenet classification with deep convolutional neural networks.
\newblock {\em Communications of the ACM}, 60(6):84--90, 2017.

\bibitem{zhao2019object}
Zhong-Qiu Zhao, Peng Zheng, Shou-tao Xu, and Xindong Wu.
\newblock Object detection with deep learning: A review.
\newblock {\em IEEE transactions on neural networks and learning systems},
  30(11):3212--3232, 2019.

\bibitem{shen2017deep}
Dinggang Shen, Guorong Wu, and Heung-Il Suk.
\newblock Deep learning in medical image analysis.
\newblock {\em Annual review of biomedical engineering}, 19:221, 2017.

\bibitem{simonyan2013deep}
Karen Simonyan, Andrea Vedaldi, and Andrew Zisserman.
\newblock Deep inside convolutional networks: Visualising image classification
  models and saliency maps.
\newblock {\em arXiv preprint arXiv:1312.6034}, 2013.

\bibitem{sundararajan2017axiomatic}
Mukund Sundararajan, Ankur Taly, and Qiqi Yan.
\newblock Axiomatic attribution for deep networks.
\newblock In {\em International conference on machine learning}, pages
  3319--3328. PMLR, 2017.

\bibitem{shrikumar2017learning}
Avanti Shrikumar, Peyton Greenside, and Anshul Kundaje.
\newblock Learning important features through propagating activation
  differences.
\newblock In {\em International conference on machine learning}, pages
  3145--3153. PMLR, 2017.

\bibitem{ghorbani2019interpretation}
Amirata Ghorbani, Abubakar Abid, and James Zou.
\newblock Interpretation of neural networks is fragile.
\newblock In {\em Proceedings of the AAAI conference on artificial
  intelligence}, volume~33, pages 3681--3688, 2019.

\bibitem{heo2019fooling}
Juyeon Heo, Sunghwan Joo, and Taesup Moon.
\newblock Fooling neural network interpretations via adversarial model
  manipulation.
\newblock {\em Advances in Neural Information Processing Systems}, 32, 2019.

\bibitem{smilkov2017smoothgrad}
Daniel Smilkov, Nikhil Thorat, Been Kim, Fernanda Vi{\'e}gas, and Martin
  Wattenberg.
\newblock Smoothgrad: removing noise by adding noise.
\newblock {\em arXiv preprint arXiv:1706.03825}, 2017.

\bibitem{zou2005regularization}
Hui Zou and Trevor Hastie.
\newblock Regularization and variable selection via the elastic net.
\newblock {\em Journal of the royal statistical society: series B (statistical
  methodology)}, 67(2):301--320, 2005.

\bibitem{meier2008group}
Lukas Meier, Sara Van De~Geer, and Peter B{\"u}hlmann.
\newblock The group lasso for logistic regression.
\newblock {\em Journal of the Royal Statistical Society: Series B (Statistical
  Methodology)}, 70(1):53--71, 2008.

\bibitem{zhou2016learning}
Bolei Zhou, Aditya Khosla, Agata Lapedriza, Aude Oliva, and Antonio Torralba.
\newblock Learning deep features for discriminative localization.
\newblock In {\em Proceedings of the IEEE conference on computer vision and
  pattern recognition}, pages 2921--2929, 2016.

\bibitem{chattopadhay2018grad}
Aditya Chattopadhay, Anirban Sarkar, Prantik Howlader, and Vineeth~N
  Balasubramanian.
\newblock Grad-cam++: Generalized gradient-based visual explanations for deep
  convolutional networks.
\newblock In {\em 2018 IEEE winter conference on applications of computer
  vision (WACV)}, pages 839--847. IEEE, 2018.

\bibitem{selvaraju2017grad}
Ramprasaath~R Selvaraju, Michael Cogswell, Abhishek Das, Ramakrishna Vedantam,
  Devi Parikh, and Dhruv Batra.
\newblock Grad-cam: Visual explanations from deep networks via gradient-based
  localization.
\newblock In {\em Proceedings of the IEEE international conference on computer
  vision}, pages 618--626, 2017.

\bibitem{rebuffi2020there}
Sylvestre-Alvise Rebuffi, Ruth Fong, Xu~Ji, and Andrea Vedaldi.
\newblock There and back again: Revisiting backpropagation saliency methods.
\newblock In {\em Proceedings of the IEEE/CVF Conference on Computer Vision and
  Pattern Recognition}, pages 8839--8848, 2020.

\bibitem{wagner2019interpretable}
Jorg Wagner, Jan~Mathias Kohler, Tobias Gindele, Leon Hetzel, Jakob~Thaddaus
  Wiedemer, and Sven Behnke.
\newblock Interpretable and fine-grained visual explanations for convolutional
  neural networks.
\newblock In {\em Proceedings of the IEEE/CVF Conference on Computer Vision and
  Pattern Recognition}, pages 9097--9107, 2019.

\bibitem{fong2017interpretable}
Ruth~C Fong and Andrea Vedaldi.
\newblock Interpretable explanations of black boxes by meaningful perturbation.
\newblock In {\em Proceedings of the IEEE international conference on computer
  vision}, pages 3429--3437, 2017.

\bibitem{lim2021building}
Dohun Lim, Hyeonseok Lee, and Sungchan Kim.
\newblock Building reliable explanations of unreliable neural networks: locally
  smoothing perspective of model interpretation.
\newblock In {\em Proceedings of the IEEE/CVF Conference on Computer Vision and
  Pattern Recognition}, pages 6468--6477, 2021.

\bibitem{dabkowski2017real}
Piotr Dabkowski and Yarin Gal.
\newblock Real time image saliency for black box classifiers.
\newblock {\em Advances in neural information processing systems}, 30, 2017.

\bibitem{ivanovs2021perturbation}
Maksims Ivanovs, Roberts Kadikis, and Kaspars Ozols.
\newblock Perturbation-based methods for explaining deep neural networks: A
  survey.
\newblock {\em Pattern Recognition Letters}, 150:228--234, 2021.

\bibitem{madry2017towards}
Aleksander Madry, Aleksandar Makelov, Ludwig Schmidt, Dimitris Tsipras, and
  Adrian Vladu.
\newblock Towards deep learning models resistant to adversarial attacks.
\newblock {\em arXiv preprint arXiv:1706.06083}, 2017.

\bibitem{dombrowski2019explanations}
Ann-Kathrin Dombrowski, Maximillian Alber, Christopher Anders, Marcel
  Ackermann, Klaus-Robert M{\"u}ller, and Pan Kessel.
\newblock Explanations can be manipulated and geometry is to blame.
\newblock {\em Advances in Neural Information Processing Systems}, 32, 2019.

\bibitem{subramanya2019fooling}
Akshayvarun Subramanya, Vipin Pillai, and Hamed Pirsiavash.
\newblock Fooling network interpretation in image classification.
\newblock In {\em Proceedings of the IEEE/CVF International Conference on
  Computer Vision}, pages 2020--2029, 2019.

\bibitem{levine2019certifiably}
Alexander Levine, Sahil Singla, and Soheil Feizi.
\newblock Certifiably robust interpretation in deep learning.
\newblock {\em arXiv preprint arXiv:1905.12105}, 2019.

\bibitem{xu2018structured}
Kaidi Xu, Sijia Liu, Pu~Zhao, Pin-Yu Chen, Huan Zhang, Quanfu Fan, Deniz
  Erdogmus, Yanzhi Wang, and Xue Lin.
\newblock Structured adversarial attack: Towards general implementation and
  better interpretability.
\newblock {\em arXiv preprint arXiv:1808.01664}, 2018.

\bibitem{krizhevsky2009learning}
Alex Krizhevsky, Geoffrey Hinton, et~al.
\newblock Learning multiple layers of features from tiny images.
\newblock 2009.

\bibitem{deng2009imagenet}
Jia Deng, Wei Dong, Richard Socher, Li-Jia Li, Kai Li, and Li~Fei-Fei.
\newblock Imagenet: A large-scale hierarchical image database.
\newblock In {\em 2009 IEEE conference on computer vision and pattern
  recognition}, pages 248--255. IEEE, 2009.

\bibitem{he2016deep}
Kaiming He, Xiangyu Zhang, Shaoqing Ren, and Jian Sun.
\newblock Deep residual learning for image recognition.
\newblock In {\em Proceedings of the IEEE conference on computer vision and
  pattern recognition}, pages 770--778, 2016.

\bibitem{tan2019efficientnet}
Mingxing Tan and Quoc Le.
\newblock Efficientnet: Rethinking model scaling for convolutional neural
  networks.
\newblock In {\em International conference on machine learning}, pages
  6105--6114. PMLR, 2019.

\bibitem{wang2004image}
Zhou Wang, Alan~C Bovik, Hamid~R Sheikh, and Eero~P Simoncelli.
\newblock Image quality assessment: from error visibility to structural
  similarity.
\newblock {\em IEEE transactions on image processing}, 13(4):600--612, 2004.

\bibitem{bertsekas1997nonlinear}
Dimitri~P Bertsekas.
\newblock Nonlinear programming.
\newblock {\em Journal of the Operational Research Society}, 48(3):334--334,
  1997.

\bibitem{zhou2018fenchel}
Xingyu Zhou.
\newblock On the fenchel duality between strong convexity and lipschitz
  continuous gradient.
\newblock {\em arXiv preprint arXiv:1803.06573}, 2018.

\bibitem{landsman2013note}
Zinoviy Landsman, Steven Vanduffel, and Jing Yao.
\newblock A note on stein's lemma for multivariate elliptical distributions.
\newblock {\em Journal of Statistical Planning and Inference},
  143(11):2016--2022, 2013.

\end{thebibliography}
}
%\clearpage
\begin{appendices}
\section{Proofs}

\subsection{Proof of Theorem \ref{Theorem: Moreau Envelope}}
Based on Theorem \ref{Theorem: Moreau Envelope}'s assumption, $g$ is $\lambda$-weakly-convex, which means that $g(\mathbf{x}) = \Phi(\mathbf{x})-\frac{\lambda}{2}\Vert \mathbf{x}\Vert^2_2$ for a convex function $\Phi:\mathbb{R}^d\rightarrow\mathbb{R}$. Therefore, we can rewrite the definition of the Moreau envelope as

\begin{align*}
    &g^{\rho} (\mathbf{x}) \\
    =\, & \min_{\widetilde{\mathbf{x}}\in\mathbb{R}^d}\:\Phi \bigl(\widetilde{\mathbf{x}}\bigr) -\frac{\lambda}{2}\bigl\Vert \widetilde{\mathbf{x}}\bigr\Vert^2_2 +\frac{1}{2\rho}\bigl\Vert \widetilde{\mathbf{x}} - \mathbf{x}\bigr\Vert_2^2 \\
    =\, & \min_{\widetilde{\mathbf{x}}\in\mathbb{R}^d}\:\Phi \bigl(\widetilde{\mathbf{x}}\bigr) +\bigl(\frac{1}{2\rho}-\frac{\lambda}{2}\bigr)\bigl\Vert \widetilde{\mathbf{x}}\bigr\Vert^2_2 -\frac{1}{\rho}\mathbf{x}^\top\widetilde{\mathbf{x}}  +\frac{1}{2\rho}\bigl\Vert \mathbf{x}\bigr\Vert_2^2.
    \label{eq:MoreauGrad}
\end{align*}
Note that based on the theorem's assumption the coefficient $\frac{1}{2\rho}-\frac{\lambda}{2} >0$ is positive. Therefore, the function $h:\mathbb{R}^d\rightarrow\mathbb{R}$ defined as
$$h(\tilde{\mathbf{x}}):= \bigl(\frac{1}{2\rho}-\frac{\lambda}{2}\bigr)\bigl\Vert \widetilde{\mathbf{x}}\bigr\Vert^2_2 -\frac{1}{\rho}\mathbf{x}^\top\widetilde{\mathbf{x}} $$ 
is a strongly-convex quadratic function with strong-convexity degree $\frac{1}{\rho} -\lambda$.  As a result, $\Phi(\tilde{\mathbf{x}})+h(\tilde{\mathbf{x}})$ will also be a $(\frac{1}{\rho}-\lambda)$-strongly convex function with a unique locally (and globally) optimal solution ${\widetilde{x}}^*_\rho(\mathbf{x})$.

Since we proved that the objective function in the Moreau envelope optimization is a strongly-convex function, we can apply the Danskin's theorem \cite{bertsekas1997nonlinear} to show that the following holds where $\psi(\mathbf{x},\tilde{\mathbf{x}}):= g(\tilde{\mathbf{x}}) + \frac{1}{2\rho}\Vert\tilde{\mathbf{x}} - \mathbf{x} \Vert^2_2$ denotes the optimization objective
\begin{align*}
    \nabla g^\rho(\mathbf{x}) \, &= \, \frac{\partial \psi}{\partial \mathbf{x}} \bigl(\mathbf{x},{\widetilde{x}}^*_\rho(\mathbf{x})\bigr) \\
    &= \frac{1}{\rho}\bigl( \mathbf{x}  - {\widetilde{x}}^*_\rho(\mathbf{x}) \bigr),
\end{align*}
which proves Part $2$ of Theorem \ref{Theorem: Moreau Envelope}. On the other hand, we showed that $\psi(\mathbf{x},\tilde{\mathbf{x}})$ is a convex function of $\tilde{\mathbf{x}}$, and therefore the first-order necessary optimality condition implies that
\begin{equation*}
    \nabla g\bigl( {\widetilde{x}}^*_\rho(\mathbf{x})\bigr) + \frac{1}{\rho}\bigl(  {\widetilde{x}}^*_\rho(\mathbf{x}) - \mathbf{x} \bigr) = \mathbf{0}.
\end{equation*}
The above identities reveal that 
\begin{equation*}
     \nabla g^\rho(\mathbf{x}) = \nabla g\bigl( {\widetilde{x}}^*_\rho(\mathbf{x})\bigr),
\end{equation*}
 which proves Part $1$ of Theorem \ref{Theorem: Moreau Envelope}. Finally, we note that as shown above the following holds

 \begin{align*}
    &\rho \, g^{\rho} (\mathbf{x}) \\
    =\, & \frac{1}{2}\bigl\Vert \mathbf{x}\bigr\Vert_2^2 + \min_{\widetilde{\mathbf{x}}\in\mathbb{R}^d}\left\{\Phi \bigl(\widetilde{\mathbf{x}}\bigr) +\frac{1-\lambda\rho}{2}\bigl\Vert \widetilde{\mathbf{x}}\bigr\Vert^2_2 -\mathbf{x}^\top\widetilde{\mathbf{x}}\right\}
    \\
    =\, & \frac{1}{2}\bigl\Vert \mathbf{x}\bigr\Vert_2^2 - \max_{\widetilde{\mathbf{x}}\in\mathbb{R}^d}\left\{\mathbf{x}^\top\widetilde{\mathbf{x}}-\Phi \bigl(\widetilde{\mathbf{x}}\bigr) -\frac{1-\lambda\rho}{2}\bigl\Vert \widetilde{\mathbf{x}}\bigr\Vert^2_2\right\}
\end{align*}
Therefore, $\rho  g^{\rho} (\mathbf{x})$ is the subtraction of the Fenchel conjugate of $s(\mathbf{x}) =\Phi \bigl(\mathbf{x}\bigr) +\frac{1-\lambda\rho}{2}\bigl\Vert \mathbf{x}\bigr\Vert^2_2$ from the $1$-strongly-convex $\frac{1}{2}\bigl\Vert \mathbf{x}\bigr\Vert_2^2$. Then, we apply the result that the Fenchel conjugate of a $\mu$-strongly convex function is $\frac{1}{\mu}$-smooth convex function \cite{zhou2018fenchel}. Therefore, the following Fenchel conjugate $$s^{\star}(\mathbf{x}):=\max_{\widetilde{\mathbf{x}}\in\mathbb{R}^d}\left\{\mathbf{x}^\top\widetilde{\mathbf{x}}-\Phi \bigl(\widetilde{\mathbf{x}}\bigr) -\frac{1-\lambda\rho}{2}\bigl\Vert \widetilde{\mathbf{x}}\bigr\Vert^2_2\right\}$$ is a $\frac{1}{1-\rho\lambda}$-smooth convex function. Since, we subtract two convex functions from each other where the second one has a constant Hessian $I$, then the resulting function will be smooth of the following degree:
\begin{equation*}
   \frac{1}{\rho}\times\max\bigl\{ \bigl\vert\frac{1}{1-\rho\lambda} -1\bigr\vert  \, ,\, \vert 0 - 1\vert \bigr\} = \max\bigl\{\frac{\lambda}{1-\rho\lambda} , \frac{1}{\rho} \bigr\},
\end{equation*}
which completes the proof of the theorem's final part.
\subsection{Proof of Proposition \ref{Prop: Gaussian smoothing}}
To prove this theorem, we apply the multivariate version of Stein's lemma in \cite{landsman2013note}, which shows that for a Lipschitz-continuous function $g:\mathbb{R}^d\rightarrow\mathbb{R}$ and Gaussian $\mathbf{Z}\sim\mathcal{N}(\mathbf{0},\sigma^2 I)$ we have
\begin{equation*}
    \mathbb{E}\bigl[ \nabla g(\mathbf{x}+\mathbf{Z}) \bigr] = \mathbb{E}\bigl[ g(\mathbf{x}+\mathbf{Z}) \frac{\mathbf{Z}}{\sigma^2} \bigr]
\end{equation*}
Then, for every $\mathbf{x},\mathbf{x}'\in\mathbb{R}^d$ we can write
\begin{align*}
    & \bigl\Vert \nabla h_{\mathbf{w},c}(\mathbf{x}) - \nabla h_{\mathbf{w},c}(\mathbf{x}') \bigr\Vert_2 \\
    =\, & \left\Vert \mathbb{E}\bigl[ \nabla f_{\mathbf{w},c}(\mathbf{x}+\mathbf{Z})\bigr]- \mathbb{E}\bigl[ \nabla f_{\mathbf{w},c}(\mathbf{x}'+\mathbf{Z})\bigr] \right\Vert_2 \\
    =\, & \left\Vert \mathbb{E}\left[ \frac{\mathbf{Z}}{\sigma^2} f_{\mathbf{w},c}(\mathbf{x}+\mathbf{Z})\right]- \mathbb{E}\left[ \frac{\mathbf{Z}}{\sigma^2} f_{\mathbf{w},c}(\mathbf{x}'+\mathbf{Z})\right] \right\Vert_2 \\
    =\, & \left\Vert \mathbb{E}\biggl[ \frac{\mathbf{Z}}{\sigma^2} \bigl( f_{\mathbf{w},c}(\mathbf{x}+\mathbf{Z})- f_{\mathbf{w},c}(\mathbf{x}'+\mathbf{Z})\bigr)\biggr] \right\Vert_2 \\
    \le\, &  \mathbb{E}\biggl[ \left\Vert\frac{\mathbf{Z}}{\sigma^2} \bigl( f_{\mathbf{w},c}(\mathbf{x}+\mathbf{Z})- f_{\mathbf{w},c}(\mathbf{x}'+\mathbf{Z})\bigr)\right\Vert_2\biggr]  \\
    =\, &  \mathbb{E}\biggl[ \frac{\Vert \mathbf{Z}\Vert_2}{\sigma^2} \bigl\vert f_{\mathbf{w},c}(\mathbf{x}+\mathbf{Z})- f_{\mathbf{w},c}(\mathbf{x}'+\mathbf{Z})\bigr\vert\biggr]  \\
    \le\, &  \mathbb{E}\biggl[ \frac{\Vert \mathbf{Z}\Vert_2}{\sigma^2} L \Vert \mathbf{x}-\mathbf{x}'\Vert_2 \biggr] \\
    =\, & \frac{L\Vert \mathbf{x}-\mathbf{x}'\Vert_2}{\sigma^2} \mathbb{E}\bigl[ {\Vert \mathbf{Z}\Vert_2}  \bigr] \\
    = \, & \frac{L\Vert \mathbf{x}-\mathbf{x}'\Vert_2}{\sigma^2} \sqrt{\frac{2d\sigma^2}{\pi}} \\
    \le \, & \frac{L\sqrt{d}\Vert \mathbf{x}-\mathbf{x}'\Vert_2}{\sigma}.
\end{align*}
Therefore, the gradient of $h_{\mathbf{w},c}$ will be $\frac{L\sqrt{d}}{\sigma}$-Lipschitz, which means that $h_{\mathbf{w},c}$ is $\frac{L\sqrt{d}}{\sigma}$-smooth and satisfies the following inequality for every $\mathbf{x},\mathbf{x}'$:
\begin{equation*}
    \biggl\vert h_{\mathbf{w},c}(\mathbf{x}') - \nabla h_{\mathbf{w},c}(\mathbf{x})^\top (\mathbf{x}'-\mathbf{x}) \biggr\vert \le \frac{L\sqrt{d}}{2\sigma} \Vert \mathbf{x} -\mathbf{x}' \Vert^2_2.
\end{equation*}
As a result, $h_{\mathbf{w},c}$ will be $\frac{L\sqrt{d}}{\sigma}$-weakly-convex, and the proof is complete.
\subsection{Proof of Theorem \ref{Thm: Sparse MoreauGrad} }
To prove Theorem \ref{Thm: Sparse MoreauGrad}, we note that the additional $L_1$-norm-based and $L_{2,1}$-norm-based functions $\eta\Vert \widetilde{\mathbf{x}} - \mathbf{x}\Vert_1$, $\eta\Vert \widetilde{\mathbf{x}} - \mathbf{x}\Vert_{2,1}$  are both convex functions of $\widetilde{\mathbf{x}}$. Therefore, repeating the proof of Theorem \ref{Theorem: Moreau Envelope}, we can show the objective function of both $L_1$-Moreau envelope and $L_{2,1}$-Moreau envelope are both strongly-convex functions of strong-convexity degree $\frac{1}{\rho} -\lambda$. Therefore, the optimization of $L_1$-Moreau envelope and $L_{2,1}$-Moreau envelope has a unique locally and globally optimal solution. Also, applying a change of variable $\boldsymbol{\delta}:= \widetilde{\mathbf{x}} - \mathbf{x}$ gives us the following formulations of the $L_1$-Moreau envelope and $L_{2,1}$-Moreau envelope:
 \begin{align*}
        g_{L_1}^{\rho,\eta}(\mathbf{x})\, :=& \, \min_{\boldsymbol{\delta}\in\mathbb{R}^d}\: g({\mathbf{x}}+ \boldsymbol{\delta}) + \frac{1}{2\rho} \bigl\Vert\boldsymbol{\delta} \bigr\Vert^2_2 + {\eta} \bigl\Vert\boldsymbol{\delta} \bigr\Vert_1, \\
        g_{L_{2,1}}^{\rho,\eta}(\mathbf{x})\, :=& \, \min_{\boldsymbol{\delta}\in\mathbb{R}^d}\: g({\mathbf{x}}+\boldsymbol{\delta}) + \frac{1}{2\rho} \bigl\Vert\boldsymbol{\delta} \bigr\Vert^2_2 + {\eta} \bigl\Vert\boldsymbol{\delta} \bigr\Vert_{2,1}.
    \end{align*}
  Then, applying the Danskin's theorem \cite{bertsekas1997nonlinear} will complete the proof of Theorem \ref{Thm: Sparse MoreauGrad}'s part 1.

Next, we define the proximal operator of the $L_1$-norm and $L_{2,1}$-norm functions as
\begin{align*}
    \text{\rm prox}_{\eta \Vert\cdot\Vert_1}(\mathbf{x}) :=& \underset{\mathbf{x}'\in\mathbb{R}^d}{\arg\!\min}\: \eta \Vert\mathbf{x}'\Vert_1 + \frac{1}{2}\bigl\Vert \mathbf{x}' - \mathbf{x}\bigr\Vert^2_2 \\
    \text{\rm prox}_{\eta \Vert\cdot\Vert_{2,1}}(\mathbf{x}) :=& \underset{\mathbf{x}'\in\mathbb{R}^d}{\arg\!\min}\: \eta \Vert\mathbf{x}'\Vert_{2,1} + \frac{1}{2}\bigl\Vert \mathbf{x}' - \mathbf{x}\bigr\Vert^2_2.
\end{align*}
As well-known in the optimization literature, the proximal operator of the $L_1$ and $L_{2,1}$ norms reduce to the soft-thresholding functions defined in the main text as:
\begin{align*}
    \text{\rm prox}_{\eta \Vert\cdot\Vert_1}(\mathbf{x}) =& \text{\rm ST}_{\eta}(\mathbf{x}) \\
    \text{\rm prox}_{\eta \Vert\cdot\Vert_{2,1}}(\mathbf{x}) =& \text{\rm GST}_{\eta}(\mathbf{x}).
\end{align*}
Then, since the objective functions of the $L_1$ and $L_{2,1}$ Moreau envelope are the summation of the following two convex functions (w.r.t. $\boldsymbol{\delta}$) $h_\mathbf{x}(\boldsymbol{\delta}):= g({\mathbf{x}}+ \boldsymbol{\delta}) + \frac{1}{2\rho} \bigl\Vert\boldsymbol{\delta} \bigr\Vert^2_2$ and $t(\boldsymbol{\delta})=\eta \Vert \boldsymbol{\delta}\Vert$, the optimal solution $\boldsymbol{\delta}^*$ will satisfy the following equation for every $\gamma >0$:
\begin{equation*}
    \boldsymbol{\delta}^* = \text{\rm prox}_{\gamma \eta\Vert\cdot \Vert}\bigl(\boldsymbol{\delta}^* - \gamma\nabla h_\mathbf{x}(\boldsymbol{\delta}^*)\bigr).
\end{equation*}
If we choose $\gamma = \rho$, the above will reduce to 
\begin{equation*}
    \boldsymbol{\delta}^* = \text{\rm prox}_{\rho \eta\Vert\cdot \Vert}\bigl(-\rho\nabla g(\mathbf{x}+\boldsymbol{\delta}^*)\bigr).
\end{equation*}
This identity proves Part 2 of Theorem \ref{Thm: Sparse MoreauGrad}. Furthermore, note that the above implies that
\begin{equation*}
    (\mathbf{x}+\boldsymbol{\delta}^*) - \text{\rm prox}_{\rho \eta\Vert\cdot \Vert}\bigl(-\rho\nabla g(\mathbf{x}+\boldsymbol{\delta}^*)\bigr) \, = \, \mathbf{x}.
\end{equation*}
In other words, if we use $\text{\rm Id}$ to denote the identity map we will get:
\begin{equation*}
    \boldsymbol{\delta}^*(\mathbf{x}) = \biggl(\bigl(\operatorname{Id} + \text{\rm prox}_{\rho \eta\Vert\cdot \Vert}\circ \rho  \nabla g\bigr)^{-1}-\operatorname{Id}\biggr) (\mathbf{x})
\end{equation*}
Note that in the above $\operatorname{Id} + \text{\rm prox}_{\rho \eta\Vert\cdot \Vert}\circ \rho  \nabla g$  will be a $(1-\rho\lambda)$-monotone operator, where we call $h:\mathbb{R}^d\rightarrow \mathbb{R}^d$ $\tau$-monotone if for every $\mathbf{x,y}\in\mathbb{R}^d$:
\begin{equation*}
    \bigl(\mathbf{y} - \mathbf{x}\bigr)^\top\bigl( h(\mathbf{y}) -h(\mathbf{x})\bigr) \ge \tau \Vert \mathbf{y} - \mathbf{x}\Vert^2_2.
\end{equation*}
The monotonicity is due to the fact that the gradient of a $\lambda$-weakly convex function can be seen to be $-\lambda$-monotone and the proximal operator is known to be $1$-monotone. Hence, $ \boldsymbol{\delta}^*(\mathbf{x})$ will be a Lipschitz function with the following Lipschitz constant (note that $\bigl(\operatorname{Id} + \text{\rm prox}_{\rho \eta\Vert\cdot \Vert}\circ \rho  \nabla g\bigr)^{-1}$ is a monotone function with a degree between $0$ and $\frac{1}{1-\rho\lambda}$):
\begin{equation*}
   \max\bigl\{ \bigl\vert\frac{1}{1-\rho\lambda} -1\bigr\vert  \, ,\, \vert 0 - 1\vert \bigr\} = \max\bigl\{\frac{\rho\lambda}{1-\rho\lambda} , 1 \bigr\}.
\end{equation*}
Therefore, the $L_1$ Sparse and $L_{2,1}$ Group-sparse MoreauGrad interpretation will be $L$-Lipschitz with the following constant which is the same as the constant for MoreauGrad:
\begin{align*}
   \frac{1}{\rho}\max\bigl\{\frac{\lambda}{1-\rho\lambda} , 1 \bigr\} &= \max\bigl\{\frac{\lambda}{1-\rho\lambda} , \frac{1}{\rho} \bigr\} \\
   &= \frac{1}{\min\bigl\{\rho,\frac{1}{\lambda}- \rho\bigr\}}. 
\end{align*}
The theorem and its corollary's proofs are therefore complete.

% \begin{figure*}
%     \centering
%     \includegraphics[width=\textwidth]{Figures/groupl1_2rows.pdf}
%     \caption{Visualization of Group-Sparse MoreauGrad maps with different coefficient $\eta$'s. $\eta=0$ reduces to the Vanilla MoreauGrad.}
%     \label{fig:groupnorm}
% \end{figure*}

\section{Details of Numerical Experiments and Additional Numerical Results}
\subsection{Implementation Details}
Baselines' parameter selections are based on their public official implementations or suggestions in the paper. %In SmoothGrad experiments, we follow the SmoothGrad implementation, calculating the saliency map expectation over Gaussian-perturbed samples. The Gaussian noise is independent and identically distributed as $\mathcal{N}(0, \sigma^2)$ where $\sigma = 0.1*(x_{max}-x_{min})$ and $x$ is pixel intensities. For integrated gradients method, we use their official implementation, which integrated 50 saliency maps with an all-black reference image. For DeepLIFT experiments, we use the Captum library in PyTorch. and the official implementation of Sparsified SmoothGrad with degree of sparsification $\tau=0.1$.
For experiments regarding MoreauGrad, we select $\lambda = 1$ in the Vanilla MoreauGrad, $\lambda=1, \eta=0.005$ in the Sparse MoreauGrad. For Group-Sparse MoreauGrad, we choose $\lambda=1, \eta = 0.05$, and divides the image into $16 \times 16$ non-overlapping pixel groups. The experiments are conducted in PyTorch platform with a single RTX 3090 Ti with 24 GB VRAM.

\subsection{Interpretation Attack Method}

\begin{figure}[t]
    \centering
    \includegraphics[width=\textwidth]{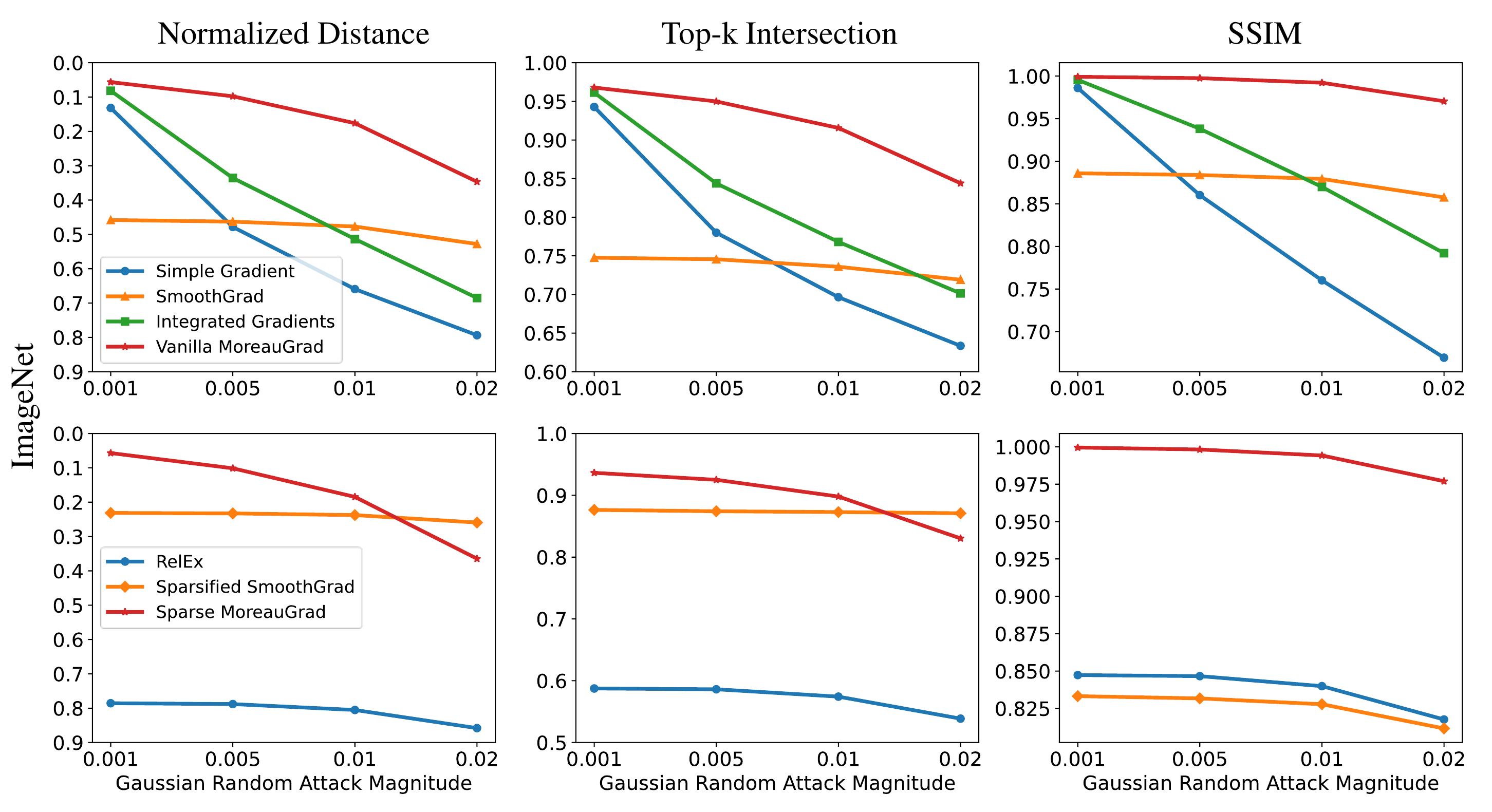}
    \caption{Quantitative robustness comparison between MoreauGrad and the baselines against Gaussian random attacks.}
    \vspace{-2mm}
    \label{fig:line_g}
\end{figure}

We adopt a gradient-based interpretation attack method which is commonly used in \cite{levine2019certifiably, ghorbani2019interpretation}. This $L_2$-bounded first-order attack method focus on attacking the most important pixels that have highest intensities in interpretation maps without changing the predictions of the neural network, which is defined as:

\begin{align}
    \mathop{\mathrm{argmin}}_\delta \; &\sum_{i \in K} m_i \\
    \text{\rm s.t.} \; \; & \Vert \delta \Vert_2 \leq \epsilon \label{eq:restriction} \\ 
    \text{\rm Prediction}(\mathbf{x}) &= \text{\rm Prediction}(\mathbf{x}+\delta)
    \label{eq:attack}
\end{align}
where $m$ denotes pixel intensity, $K$ contains most salient pixels in the interpretation map. \cref{eq:restriction} ensures the attack level is bounded, and \cref{eq:attack} controls the attack process not to change the prediction.

We also include Gaussian-random attack to examine the robustness of interpretation methods. This attack method adds a randomly generated Gaussian noise to the original input. The results are shown in \cref{fig:line_g}.

\end{appendices}
\end{document}